%% file: crvae.tex
\documentclass{article}

\usepackage[final]{neurips_2021}

\usepackage[utf8]{inputenc} 
\usepackage[T1]{fontenc}    
\usepackage{url}            
\usepackage{booktabs}       
\usepackage{amsfonts}       
\usepackage{nicefrac}       
\usepackage{microtype}      
\usepackage{graphicx}
\usepackage{multirow}

\input{preamble/preamble}

\input{preamble/preamble_acronyms}

\input{preamble/preamble_math}

\title{Consistency Regularization\\ for Variational Auto-Encoders}

\author{%
  Samarth Sinha\\
  Vector Institute\\
  University of Toronto\\
  \And
  Adji B. Dieng \\
  Google Brain \\
  Princeton University\\
}

\begin{document}

\maketitle

\begin{abstract}

\Glspl{VAE} are a powerful approach to unsupervised learning. 
They enable scalable approximate posterior inference in latent-variable models 
using \gls{VI}. A \gls{VAE} posits a variational family parameterized by a 
deep neural network---called an \emph{encoder}---that takes data as input. 
This encoder is shared across all the observations, which amortizes the cost of inference. 
However the encoder of a \gls{VAE} has the undesirable property that it maps a given 
observation and a semantics-preserving transformation of it to different latent representations. 
This ``inconsistency" of the encoder lowers the quality of the learned representations, especially for downstream tasks, and also negatively affects generalization. In this paper, we propose a regularization method to enforce 
consistency in \glspl{VAE}. The idea is to minimize the \gls{KL} divergence between 
the variational distribution when conditioning on the observation and the variational distribution when conditioning on a random semantic-preserving transformation of this observation. This regularization is applicable to any \gls{VAE}. In our experiments we apply it to four different \gls{VAE} variants on several benchmark datasets and found it always improves the quality of the learned representations but also leads to better generalization. In particular, when applied to the \gls{NVAE}, our regularization method yields state-of-the-art performance on \textsc{mnist},  \textsc{cifar}-10, and \textsc{celeba}. We also applied our method to 3D data and found it learns representations of superior quality as measured by accuracy on a downstream classification task. Finally, we show our method can even outperform the triplet loss, an advanced and popular contrastive learning-based method for representation learning. \footnote{Code for this work can be found at \url{https://github.com/sinhasam/CRVAE}}
\end{abstract}

\input{sec_intro}
\input{sec_method}
\input{sec_related}

\input{sec_experiments}
\input{sec_discussion}

\section*{Broader Impact}
\label{sec:impact}

In this paper, we propose a simple method that performs a KL-based consistency regularization scheme using data augmentation for \glspl{VAE}. The broader impact of the study includes practical applications such as graphics and computer vision applications. The method we propose improves the learned representations of \glspl{VAE}, and as an artifact, also improves their generalization to unseen data. In this regard, any implications of \glspl{VAE} also apply to this work. For example, the generative model fit by a \gls{VAE} may be used to generate artificial data such as images, text, and 3D objects. Biases may arise as a result of poor data selection. Furthermore, text generated from generative systems may amplify harmful speech contained in the data. However, the method we propose can also improve the performance of \glspl{VAE} when used in certain practical domains as we discussed in the introduction of the paper.

\section{Acknowledgements}
We thank Kevin Murphy, Ben Poole, and Augustus Odena for their comments on this work.

\nocite{jun2020distribution}
\nocite{osada2020regularization}
\nocite{hadjeres2017glsr}

{\small
\bibliography{crvae}
\bibliographystyle{icml2020}
}

\end{document}

%% file: preamble/preamble.tex
\usepackage[T1]{fontenc}
\usepackage{tgtermes}

\usepackage{amssymb}
\newcommand{\mathbold}[1]{\ensuremath{\boldsymbol{\mathbf{#1}}}}

\usepackage{subfigure}
\usepackage{scalefnt,letltxmacro}
\LetLtxMacro{\oldtextsc}{\textsc}
\renewcommand{\textsc}[1]{\oldtextsc{\scalefont{1.10}#1}}
\usepackage[ttdefault=true]{AnonymousPro}

\usepackage{caption} 
\usepackage{amsmath}

\usepackage[english]{babel}
\usepackage[parfill]{parskip}
\usepackage{afterpage}
\usepackage{enumitem}
\usepackage{framed}
\usepackage{xspace}

\usepackage{lineno}

\usepackage{ragged2e}

\newcounter{parcount}

\DeclareRobustCommand{\parhead}[1]{\textbf{#1}~}

\usepackage[usenames,dvipsnames]{xcolor}
\definecolor{shadecolor}{gray}{0.9}

\usepackage{graphicx}
\usepackage[labelfont=bf]{caption}

\usepackage{booktabs, array}

\usepackage[algoruled]{algorithm2e}
\usepackage{listings}
\usepackage{fancyvrb}
\fvset{fontsize=\small}

\usepackage[colorlinks,linktoc=all]{hyperref}
\usepackage[all]{hypcap}
\hypersetup{citecolor=MidnightBlue}
\hypersetup{linkcolor=MidnightBlue}
\hypersetup{urlcolor=MidnightBlue}
\usepackage[nameinlink]{cleveref}
\creflabelformat{equation}{#2\textup{#1}#3}  

\usepackage
[acronym,smallcaps,nowarn,section,nogroupskip,nonumberlist]{glossaries}
\glsdisablehyper{}

\lstdefinestyle{mystyle}{
    commentstyle=\color{OliveGreen},
    numberstyle=\tiny\color{black!60},
    stringstyle=\color{BrickRed},
    basicstyle=\ttfamily\scriptsize,
    breakatwhitespace=false,
    breaklines=true,
    captionpos=b,
    keepspaces=true,
    numbers=none,
    numbersep=5pt,
    showspaces=false,
    showstringspaces=false,
    showtabs=false,
    tabsize=2
}


%% file: preamble/preamble_acronyms.tex
\newacronym{CUBO}{cubo}{$\chi$ upper bound}
\newacronym{BBVI}{bbvi}{black box variational inference}
\newacronym{ELBO}{elbo}{evidence lower bound}
\newacronym{EP}{ep}{expectation propagation}
\newacronym{GP}{gp}{Gaussian process}
\newacronym{VI}{vi}{variational inference}
\newacronym{KL}{kl}{Kullback-Leibler}
\newacronym{MCMC}{mcmc}{Markov chain Monte Carlo}
\newacronym{HMC}{hmc}{Hamiltonian Monte Carlo}
\newacronym{chiVI}{chivi}{$\chi$-divergence variational inference}
\newacronym{klVI}{klvi}{$\operatorname{KL}(q || p)$ variational inference}
\newacronym{VAE}{vae}{variational auto-encoder}
\newacronym{CR-VAE}{cr-vae}{consistency-regularized variational auto-encoder}
\newacronym{IWAE}{iwae}{importance-weighted auto-encoder}
\newacronym{CR-IWAE}{cr-iwae}{consistency-regularized importance-weighted auto-encoder}
\newacronym{GAN}{gan}{generative adversarial network}
\newacronym{DAE}{dae}{denoising auto-encoder}
\newacronym{CAE}{cae}{contractive auto-encoder}
\newacronym{NVAE}{nvae}{nouveau variational auto-encoder}
\newacronym{CR-NVAE}{cr-nvae}{consistency-regularized nouveau variational auto-encoder}

%% file: preamble/preamble_math.tex
\newcommand{\mba}{\mathbold{a}}
\newcommand{\mbb}{\mathbold{b}}

\newcommand{\mbu}{\mathbold{u}}

\newcommand{\mbx}{\mathbold{x}}

\newcommand{\mbz}{\mathbold{z}}

\newcommand{\mbI}{\mathbold{I}}

\newcommand{\mbQ}{\mathbold{Q}}

\newcommand{\mbW}{\mathbold{W}}

\newcommand{\mbepsilon}{\mathbold{\epsilon}}

\newcommand{\mbeta}{\mathbold{\eta}}

\newcommand{\mbmu}{\mathbold{\mu}}

\newcommand{\mbsigma}{\mathbold{\sigma}}

%% file: sec_intro.tex

\section{Introduction}
\label{sec:introduction}
\glsresetall
\Glspl{VAE} have significantly impacted research on unsupervised learning. 
They have been used in several areas, including density estimation~\citep{vae, rezende2014stochastic}, 
image generation~\citep{gregor2015draw}, text generation~\citep{bowman2015generating, fang2019implicit}, music generation~\citep{musicVAE}, topic modeling~\citep{miao2016neural, dieng2019topic}, 
and recommendation systems~\citep{liang2018variational}. \Glspl{VAE} have also been used for different representation learning problems such as semi-supervised learning \citep{kingma2014semi}, anomaly detection \citep{an2015variational, zimmerer2018context}, language modeling \cite{bowman2015generating}, active learning \citep{vaal}, continual learning \citep{achille2018life}, and motion prediction of agents \citep{walker2016uncertain}. This widespread application of \gls{VAE} representations makes it critical that we focus on improving them.

\glspl{VAE} extend deterministic auto-encoders to probabilistic generative modeling. The encoder of a \gls{VAE} parameterizes an approximate posterior distribution over 
latent variables of a generative model. The encoder is shared between all observations,  
which amortizes the cost of posterior inference. Once fitted, the encoder of a \gls{VAE} 
can be used to obtain low-dimensional representations of data, (e.g. for downstream tasks.)  
The quality of these representations is therefore very important to a successful application 
of \glspl{VAE}. 

Researchers have looked at ways to improve the quality of the latent representations of \glspl{VAE}, 
often tackling the so-called \emph{latent variable collapse} problem---in which the approximate 
posterior distribution induced by the encoder collapses to the prior over the latent 
variables~\citep{bowman2015generating, kim2018semi, skipvae, he2019lagging, fu2019cyclical}. 

In this paper, we focus on a different problem pertaining to the latent representations of \glspl{VAE} for image data. Indeed, the encoder of a fitted \gls{VAE} tends to map 
an image and a semantics-preserving transformation of that image to different parts in the latent space. This ``inconsistency" of the encoder affects the quality of the learned representations and generalization. We propose a method to enforce consistency in \glspl{VAE}. The idea is simple and consists in maximizing the likelihood of the images while minimizing the \gls{KL} divergence between the approximate posterior distribution induced by the encoder when conditioning on the image, on one hand, and its transformation, on the other hand. This regularization technique can be applied to any \gls{VAE} variant to improve the quality of the learned representations and boost generalization performance. We call a \gls{VAE} with this form of regularization, a \gls{CR-VAE}. 

\begin{figure}
    \centering
    \subfigure[]{\includegraphics[width=0.32\textwidth]{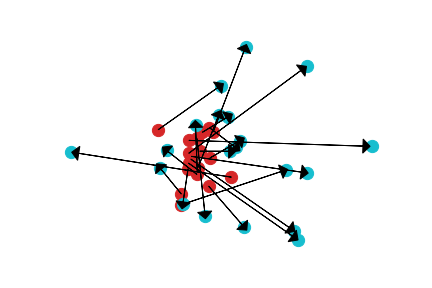}} 
    \subfigure[]{\includegraphics[width=0.32\textwidth]{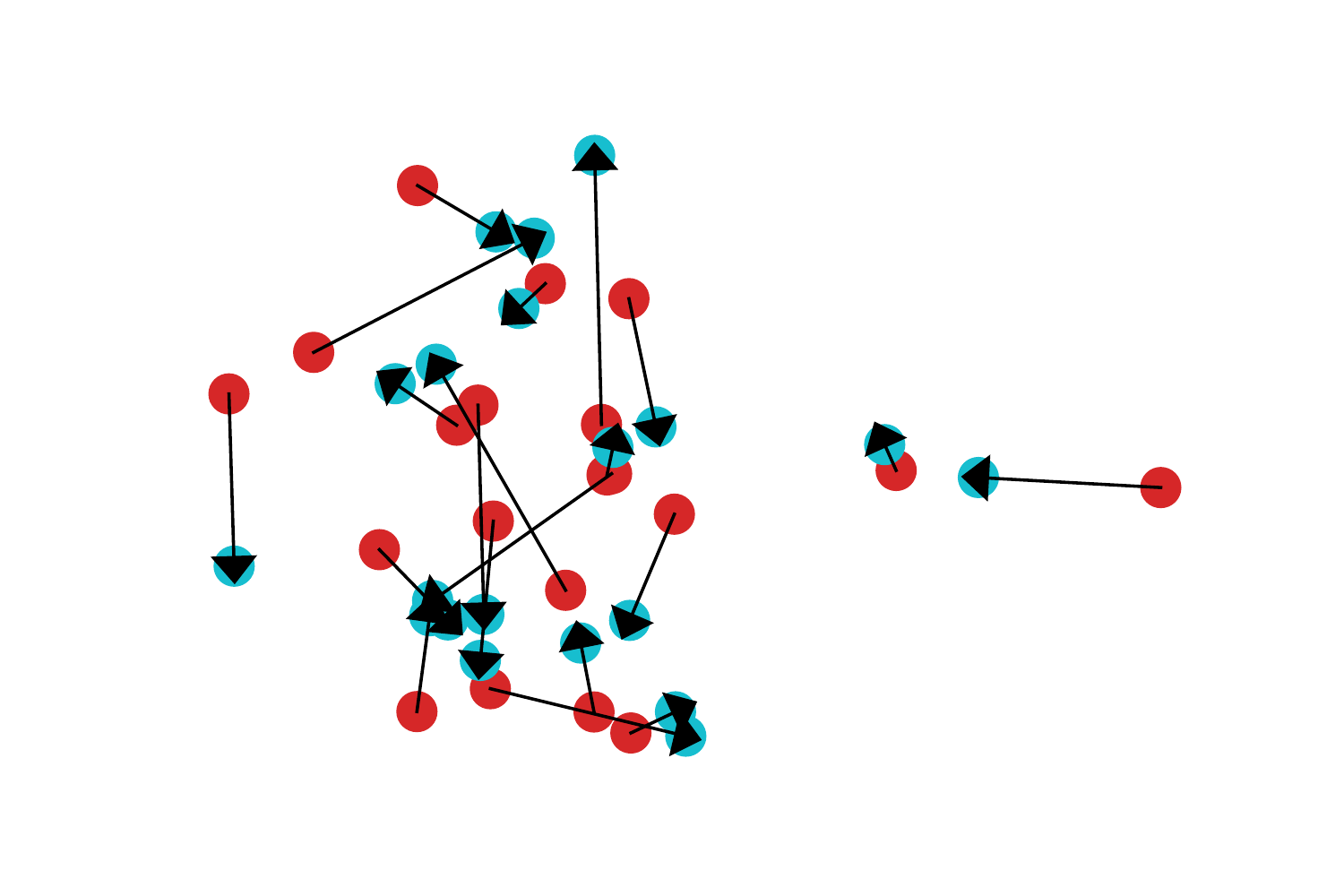}} 
    \subfigure[]{\includegraphics[width=0.32\textwidth]{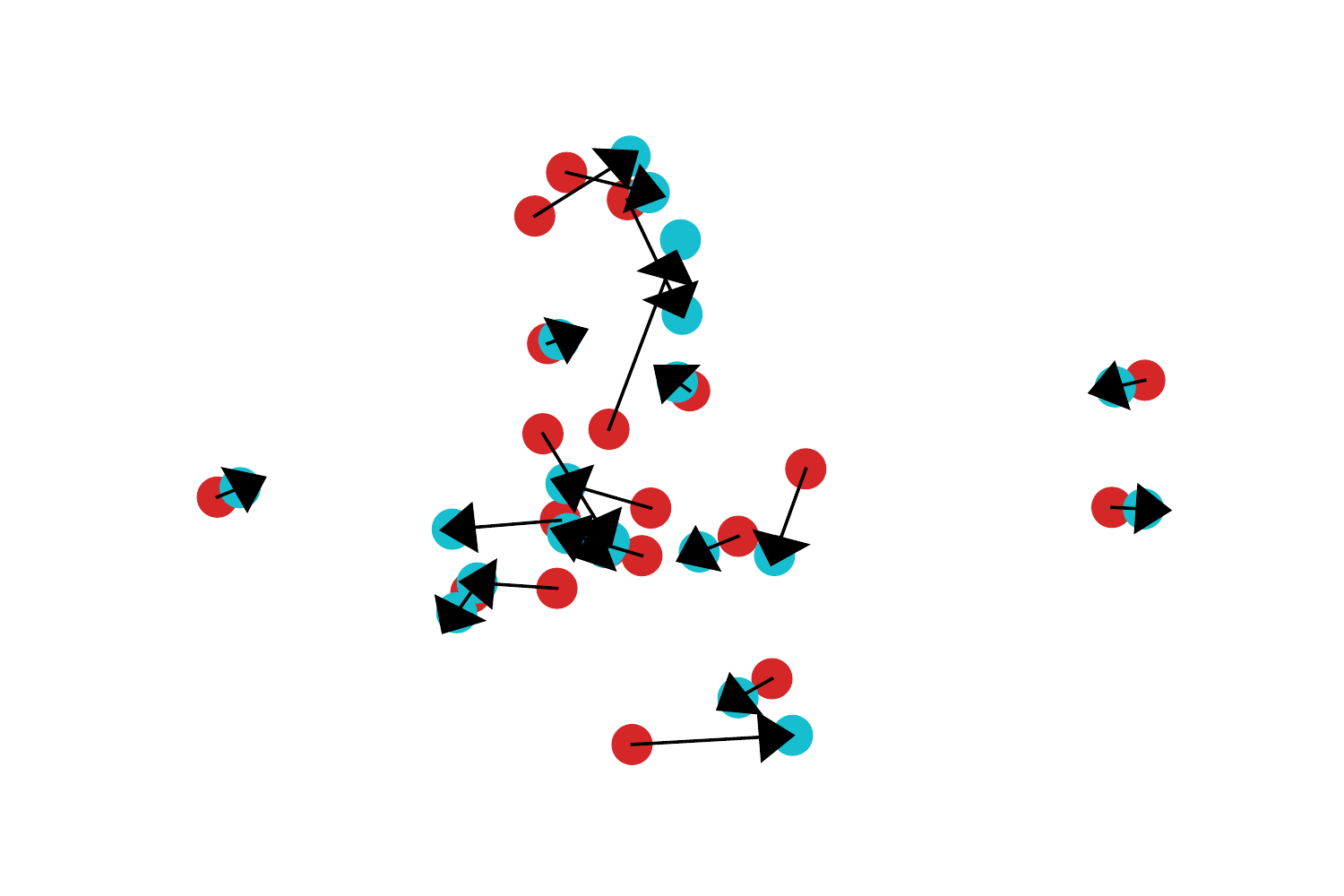}}
    \caption{Illustration of the \textit{inconsistency} problem in \glspl{VAE} and how \glspl{CR-VAE} address this problem. The \textbf{\textcolor{red}{red}} dots correspond to the representations of few images from \textsc{mnist}. The \textbf{\textcolor{cyan}{blue}} dots correspond to the representations of the transformed images. The transformations used here are rotations, translations, and scaling; they are semantics-preserving. The arrows connect the representations of any two pairs of an image and its transformation. The shorter the arrow, the better. 
    \textbf{(a)}: The \gls{VAE} maps the two sets of images to different areas in the latent space.
    \textbf{(b)}: Even when trained with the original dataset augmented with the transformed images, the \gls{VAE} still maps the two sets of images to different parts in the latent space.
    \textbf{(c)}: The \gls{CR-VAE} maps an image and its transformation to nearby areas in the latent space.}
    \label{fig:figure1}
\end{figure}

\Cref{fig:figure1} illustrates the inconsistency problem of \glspl{VAE} and how \glspl{CR-VAE} address this problem on \textsc{mnist}. The red dots are representations of a few images and the blue dots are the representations of their transformations. We applied semantics-preserving transformations: rotation, translation, and scaling. 
The \gls{VAE} maps each image and its transformation to different parts in the latent space as evidenced by the long arrows connecting each pair (a). Even when we include the transformed images to the data and fit the \gls{VAE} the inconsistency problem still occurs (b). The \gls{CR-VAE} does not suffer from the inconsistency problem; it maps each image and its transformation to nearby areas in the latent space, as evidenced by the short arrows connecting each pair (c).

In our experiments (see \Cref{sec:empirical}), we apply the proposed technique to four \gls{VAE} variants, 
the original \gls{VAE}~\citep{vae}, the \gls{IWAE}~\citep{iwae}, the $\beta$-\gls{VAE}~\citep{higgins2017beta}, and the \gls{NVAE}~\citep{vahdat2020nvae}. 
We found, on four different benchmark datasets, that \glspl{CR-VAE} always yield better representations and generalize better than their base \glspl{VAE}. In particular, \glspl{CR-NVAE} yield state-of-the-art performance on \textsc{mnist} and \textsc{cifar}-10.
We also applied \glspl{CR-VAE} to 3D data where these conclusions still hold. 

%% file: sec_method.tex

\section{Method}
\label{sec:method}

We consider a latent-variable model $p_{\theta}(\mbx, \mbz) = p_{\theta}(\mbx \vert \mbz) \cdot p(\mbz)$, 
where $\mbx$ denotes an observation and $\mbz$ is its associated latent variable. The marginal $p(\mbz)$ 
is a prior over the latent variable and $p_{\theta}(\mbx \vert \mbz)$ is an exponential family distribution 
whose natural parameter is a function of $\mbz$ parameterized by $\theta$, e.g. through a neural network. 
Our goal is to learn the parameters $\theta$ and a posterior distribution over the latent variables.  The approach 
of \glspl{VAE} is to maximize the \gls{ELBO}, a lower bound on the log marginal likelihood of the data, 
\begin{align}\label{eq:elbo}
	\mathcal{L}_{\gls{VAE}} &= \gls{ELBO} 
		= \mathbb{E}_{q_{\phi}(\mbz \vert \mbx)} \left[
		\log \left(\frac{p_{\theta}(\mbx, \mbz)}{q_{\phi}(\mbz \vert \mbx)}  \right)
		\right]
\end{align}
where $q_{\phi}(\mbz \vert \mbx)$ is an approximate posterior distribution over the latent variables. 
The idea of a \gls{VAE} is to let the parameters of the distribution $q_{\phi}(\mbz \vert \mbx)$ be 
given by the output of a neural network, with parameters $\phi$, that takes $\mbx$ as input. 
The parameters $\theta$ and $\phi$ are then jointly optimized by maximizing a Monte Carlo 
approximation of the \gls{ELBO} using the reparameterization trick~\citep{vae}. 

Consider a semantics-preserving transformation $t(\tilde \mbx \vert \mbx)$ of data $\mbx$ (e.g. rotation or translation for images.) A good representation learning algorithm should provide similar latent representations for $\mbx$ and $\tilde \mbx$. This is not the case for the \gls{VAE} that maximizes \Cref{eq:elbo} and its variants. 
Once fit to data, the encoder of a \gls{VAE} is unable to yield similar latent representations for a data $\mbx$ and its 
tranformation $\tilde \mbx$ (see \Cref{fig:figure1}). This is because there is nothing in \Cref{eq:elbo} that forces this desideratum. 

We now propose a regularization method that ensures \emph{consistency} of the encoder of a \gls{VAE}. We call a \gls{VAE} with such a regularization a \gls{CR-VAE}. The regularization proposed is applicable to many variants of the \gls{VAE} such as the \gls{IWAE}~\citep{iwae}, the $\beta$-\gls{VAE}~\citep{higgins2017beta}, and the \gls{NVAE}~\citep{vahdat2020nvae}. In what follows, we use the standard \gls{VAE}, the one that maximizes \Cref{eq:elbo}, as the base \gls{VAE} to regularize to illustrate the method. 

Consider an image $\mbx$. Denote by $t(\tilde \mbx \vert \mbx)$ the random process by which we generate $\tilde \mbx$, a semantics-preserving transformation of $\mbx$. We draw $\tilde \mbx$ from $t(\tilde \mbx \vert \mbx)$ as follows:
\begin{align}\label{eq:trans}
	\tilde \mbx &\sim t(\tilde \mbx \vert \mbx) \iff \epsilon \sim p(\epsilon) \text{ and } \tilde \mbx = g(\mbx, \epsilon)
	.
\end{align}
Here $g(\mbx, \epsilon)$ is a semantics-preserving transformation of the image $\mbx$, e.g. translation with random length $\epsilon$ drawn from $p(\epsilon) = \mathcal{U}[-\delta, \delta]$ for some threshold $\delta$. A \gls{CR-VAE} then maximizes 
\begin{equation}
    \label{eq:crvae}
    \begin{split}
        \mathcal{L}_{\gls{CR-VAE}}(\mbx) = \;
		 &\mathcal{L}_{\gls{VAE}}(\mbx) + \mathbb{E}_{t(\tilde \mbx \vert \mbx)}\left[\mathcal{L}_{\gls{VAE}}(\tilde\mbx)\right]- \lambda \cdot \mathcal{R}(\mbx, \phi)
    \end{split}
\end{equation}
where the regularization term $\mathcal{R}(\mbx,\phi)$ is
\begin{align}\label{eq:reg}
    \mathcal{R}(\mbx, \phi) &= \mathbb{E}_{t(\tilde \mbx \vert \mbx)} \left[\gls{KL}\left(q_{\phi}(\mbz \vert \tilde\mbx) \vert\vert q_{\phi}(\mbz \vert \mbx) \right)\right]
    .
\end{align}

Maximizing the objective in \Cref{eq:crvae} maximizes the likelihood of the data and their augmentations while enforcing consistency through $\mathcal{R}(\mbx, \phi)$. Minimizing $\mathcal{R}(\mbx,  \phi)$, which only affects the encoder (with parameters $\phi$), forces each observation and the corresponding augmentations to lie close to each other in the latent space. The hyperparameter $\lambda \geq 0$ controls the strength of this constraint. 

The objective in \Cref{eq:crvae} is intractable but we can easily approximate it using Monte Carlo with the reparameterization trick. In particular, we approximate the regularization term with one sample from $t(\tilde\mbx \vert \mbx)$ and make the dependence to this sample explicit using the notation $\mathcal{R}(\mbx,  \tilde\mbx, \phi)$. \Cref{alg:full_algo} illustrates this in greater detail. Although we show the application of consistency regularization using the \gls{VAE} that maximizes the \gls{ELBO}, $\mathcal{L}_{\gls{VAE}}(\cdot)$ in \Cref{eq:crvae} can be replaced with any \gls{VAE} objective. 

\begin{algorithm}[t]
  \SetAlgoNoLine
    \DontPrintSemicolon
    \SetKwInOut{KwInput}{input}
    \SetKwInOut{KwOutput}{output}
    \KwInput{Data $\mbx$, consistency regularization strength $\lambda$, latent space dimensionality K}
    Initialize parameters $\theta, \phi$\;
    \For{\emph{iteration} $t=1,2,\ldots$}{
      Draw minibatch of observations $\{\mbx_n\}_{n=1}^{B}$\;
      \For{ $n=1,\ldots, B$}{
      Transform the data: $ \mbepsilon_n \sim p(\mbepsilon_n)$ and $\tilde{\mbx}_n = T(\mbx_n,\mbepsilon_n)$ \;
      Get variational mean and variance for the data: \;
      $\quad \quad \mbmu_n = \mbW^\top\text{NN}(\mbx_n; \phi) + \mba$ and $\mbsigma_n = \text{softplus}(\mbQ^\top\text{NN}(\mbx_n; \phi) + \mbb)$\;
      Get $S$ samples from the variational distribution when conditioning on $\mbx_n$:\;
      $\quad \quad \mbeta^{(s)} \sim \mathcal{N}(0, \mbI) \quad \text{and} \quad \mbz^{(s)}_n = \mbmu_n + \eta^{(s)} \cdot \mbsigma_n$ for $s = 1, \dots, S$\;
      Get variational mean and variance for the transformed data:\;
      $\quad \quad \tilde{\mbmu}_n = \mbW^\top\text{NN}(\tilde{\mbx}_n; \phi) + \mba$ and $\tilde{\mbsigma}_n = \text{softplus}(\mbQ^\top\text{NN}(\tilde{\mbx}_n; \phi) + \mbb)$\;
      Get $S$ samples from the variational distribution when conditioning on $\tilde\mbx_n$:\;
      $\quad \quad \mbeta^{(s)} \sim \mathcal{N}(0, \mbI) \quad \text{and} \quad \tilde\mbz^{(s)}_n = \tilde\mbmu_n + \eta^{(s)} \cdot \tilde\mbsigma_n$ for $s = 1, \dots, S$\;
      }
      Compute $\mathcal{L}_{\gls{VAE}}(\mbx)$: \;
      $\quad \quad \mathcal{L}_{\gls{VAE}}(\mbx) \approx \frac{1}{B}\sum_{n=1}^{B}\frac{1}{S}\sum_{s=1}^{S}\left[\log p_{\theta}(\mbx_n, \mbz^{(s)}_n) - \log q_{\phi}(\mbz^{(s)}_n \vert \mbx_n) \right]$\;
      Compute $\mathcal{L}_{\gls{VAE}}(\tilde\mbx)$: \;
      $\quad \quad \mathcal{L}_{\gls{VAE}}(\tilde\mbx) \approx \frac{1}{B}\sum_{n=1}^{B}\frac{1}{S}\sum_{s=1}^{S}\left[\log p_{\theta}(\tilde\mbx_n, \tilde\mbz^{(s)}_n) - \log q_{\phi}(\tilde\mbz^{(s)}_n \vert \tilde\mbx_n) \right] $\;
      Compute KL consistency regularizer: \;
      $\quad \quad \mathcal{R}(\mbx, \tilde\mbx, \phi) = \frac{1}{2}\sum_{k=1}^{K}\left( \frac{\tilde\mbsigma^2_{nk} + (\tilde\mbmu_{nk} - \mbmu_{nk})^2}{\mbsigma^2_{nk}} -1 + 2\cdot \log \frac{\mbsigma_{nk}}{\tilde\mbsigma_{nk}}\right)$\;
      Compute final loss:\;
      $\quad \quad \mathcal{L}_{\gls{CR-VAE}}(\mbx) = \mathcal{L}_{\gls{VAE}}(\mbx) + \mathcal{L}_{\gls{VAE}}(\tilde\mbx) - \lambda \cdot \mathcal{R}(\mbx, \tilde\mbx, \phi)$\;
      Backpropagate through $\mathcal{L}(\mbx, \theta, \phi) = -\mathcal{L}_{\gls{CR-VAE}}(\mbx)$ and take a gradient step for $\theta$ and $\phi$\;
    }
    \caption{Consistency Regularization for Variational Autoencoders \label{alg:full_algo}}
 \end{algorithm}

%% file: sec_related.tex

\section{Related Work}
\label{sec:related}

Applying consistency regularization to \glspl{VAE}, as we do in this paper, has not been previously explored. 
Consistency regularization is a widely used technique for semi-supervised learning~\citep{bachman2014learning,CONSISTENCY,laine2016temporal,miyato2018virtual,UDACONSISTENCY}.
The core idea behind consistency regularization for semi-supervised learning is to force classifiers to learn representations 
that are insensitive to semantics-preserving changes to images, so as to improve classification of unlabeled images. 
Examples of semantics-preserving changes used in the literature include rotation, zoom, translation, crop, or 
adversarial attacks. Consistency is often enforced by minimizing the $\mathbb{L}_2$ distance between a classifier's 
logit output for an image and the logit output for its semantics-preserving transformation~\citep{CONSISTENCY, laine2016temporal}, 
or by minimizing the \gls{KL} divergence between the classifier's label distribution induced 
by the image and that of its tranformation~\citep{miyato2018virtual, UDACONSISTENCY}. 

More recently, consistency regularization has been applied to \glspl{GAN}~\citep{gan}.
Indeed \citet{improvingimproving} and \citet{CRGAN} show that applying consistency regularization on the
discriminator of a \gls{GAN}---also a classifier---can substantially improve its performance. 

The idea we develop in this paper differs from the works above in two ways. First, it applies consistency regularization to \glspl{VAE} for image data. 
Second, it leverages consistency regularization, not in the label or logit space, as done in the works mentioned above, 
but in the latent space. 

Although different, consistency regularization for \glspl{VAE} relates to works that study ways to constrain the 
sensitivity of encoders to various perturbations. For example, \glspl{DAE} and their variants~\citep{DAE, SDAE} corrupt 
an image $\mbx$ into $\mbx'$, typically using Gaussian noise, and then minimize the distance between the 
reconstruction of $\mbx'$ and the \textit{un-corrupted} image $\mbx$. 
The motivation is to learn representations that are insensitive to the added noise. 
Our work differs in that we do not constrain the decoder to recover the original image from the corrupted image 
but, rather, to constrain the encoder to recover the latent representation of the original image from the corrupted image 
via a \gls{KL} divergence minimization constraint. 

\Glspl{CAE}~\citep{CAE} share a similar goal with \glspl{CR-VAE}. A \gls{CAE} is an auto-encoder whose encoder is constrained by minimizing the norm of the Jacobian of the output of the encoder with respect to the input image. This norm constraint on the Jacobian forces the representations learned by the encoder to be insensitive to changes in the input. Our work differs in several main ways. First, \glspl{CR-VAE} are not deterministic auto-encoders, contrary to \glspl{CAE}. We can easily sample from a \gls{CR-VAE}, as for any \gls{VAE}, which is not the case for a \gls{CAE}. Second, a \gls{CAE} does not apply transformations to the input image, which limits the sensitivities it can learn to limit to those exhibited in the training set. Finally, \glspl{CAE} use the Jacobian to impose a consistency constraint, which are not as easy to compute as the \gls{KL} divergence we use on the variational distribution induced by the encoder.

%% file: sec_experiments.tex

\section{Empirical Study}
\label{sec:empirical}

In this section we show that a \gls{CR-VAE} improves the learned representations of its base \gls{VAE} and positively affects generalization performance 
We also show that the proposed regularization method is amenable to different \gls{VAE} variants by applying it not only to the original \gls{VAE} but also to the \gls{IWAE},  the $\beta$-\gls{VAE}, and the \gls{NVAE}. We showcase the importance of the KL regularization term by conducting an ablation study. 
We found that only regularizing with data augmentation improves performance but that accounting for the \gls{KL} term ($\lambda > 0$) further improves the quality of the learned representations and generalization.

We will conduct three sets of experiments. In the first experiment, we will apply the regularization method proposed in this paper to standard \glspl{VAE} such as the original \gls{VAE}, the \gls{IWAE}, and the $\beta$-\gls{VAE}. We use \textsc{mnist}, \textsc{omniglot}, and \textsc{celeba} as datasets for this experiment. For \textsc{celeba}, we choose the $32$x$32$ resolution for this experiment. Our results show that adding consistency regularization always improves upon the base \gls{VAE}, both in terms of the quality of the learned representations and generalization. We conduct an ablation study and also report performance of the different \gls{VAE} variants above when they are fitted with the original data and their augmentations. The results from this ablation highlight the importance of setting $\lambda > 0$. 

\begin{table}
\centering
\caption{\glspl{CR-VAE} learn better representations than their base \glspl{VAE} on all three benchmark datasets. Although fitting the base \gls{VAE} with augmentations does improve the representations, adding the consistency regularization further improves the quality of these learned representations. The value of $\beta$ for the $\beta$-\gls{VAE} is inside the parentheses.}
  \begin{tabular}{l|cccccc}
    \hline
     & \multicolumn{2}{c}{\textsc{mnist}} & \multicolumn{2}{c}{\textsc{omniglot}} & \multicolumn{2}{c}{\textsc{celeba}} \\
    \cline{2-7}
    \textbf{Method} & MI & AU & MI & AU & MI & AU \\
    \hline
    \gls{VAE} & $124.5\pm 1.1$ & $36\pm 0.8$ & $105.4\pm 1.2$  & $50\pm 0.0$ & $33.8\pm 0.2$  & $32\pm 0.9$\\
    \gls{VAE} + Aug & $125.9\pm 0.2$ & $42\pm 0.5$ & $105.9\pm 0.7$  & $50\pm 0.0$ & $34.1\pm 0.8$ & $33\pm 0.9$\\ 
    \gls{CR-VAE} & $\textbf{126.3}\pm \textbf{0.9}$ & $\textbf{47}\pm\textbf{0.5}$ & $\textbf{107.8}\pm \textbf{1.1}$  & $50\pm 0.0$ & $\textbf{34.9}\pm \textbf{0.5}$ & $\textbf{33}\pm \textbf{1.2}$\\ \hline
    \gls{IWAE} & $127.1\pm 0.7$ & $39\pm 0.5$ & $110.3\pm 1.1$ & $50\pm 0.0$ & $36.9\pm 0.5$ & $36\pm 1.6$\\
    \gls{IWAE}+Aug & $129.0\pm 0.9$ & $45\pm 0.8$ & $112.9\pm 0.7$ & $50\pm 0.0$ & $37.0\pm 0.2$ & $36\pm 1.2$\\
    \acrshort{CR-IWAE} & $\textbf{129.7}\pm \textbf{1.0}$ & $\textbf{50}\pm \textbf{0.0}$ & $\textbf{115.3}\pm \textbf{0.8}$ & $50\pm 0.0$ & $\textbf{38.4}\pm \textbf{0.5}$ & $\textbf{36}\pm \textbf{1.9}$\\\hline
    $\beta$-\gls{VAE} ($0.5$) & $284.3\pm 1.1$ & $50\pm 0.0$ & $143.4\pm 1.0$ & $50\pm 0.0$ & $75.8\pm 0.5$ & $49\pm 0.5$\\
    $\beta$-\gls{VAE} ($0.5$) + Aug & $289.3\pm 1.0$ & $50\pm 0.0$ & $159.6\pm 1.3$ & $50\pm 0.0$ & $75.7\pm 0.3$ & $49\pm 0.0$\\
    $\beta$-\gls{CR-VAE} ($0.5$) & $\textbf{291.9}\pm \textbf{0.7}$ & $50\pm 0.0$ & $\textbf{169.5}\pm \textbf{0.5}$ & $50\pm 0.0$ & $\textbf{77.1}\pm \textbf{0.1}$ & $\textbf{50}\pm \textbf{0.0}$\\\hline
     $\beta$-\gls{VAE} ($10$) & $6.3\pm 0.6$ & $8\pm 1.7$ & $1.4\pm 0.2$ & $4\pm 0.9$ & $3.6\pm 0.3$ & $7\pm 0.8$\\
    $\beta$-\gls{VAE} ($10$) + Aug & $6.5\pm 0.5$ & $9\pm 1.1$ & $1.6\pm 0.2$ & $4\pm 0.5$ & $3.7\pm 0.1$ & $7\pm 0.0$\\
    $\beta$-\gls{CR-VAE} ($10$) & $\textbf{6.9}\pm \textbf{0.6}$ & $\textbf{10}\pm \textbf{0.5}$ & $\textbf{1.6}\pm \textbf{0.1}$ & $\textbf{4}\pm \textbf{0.5}$ & $\textbf{3.7}\pm \textbf{0.4}$ & $\textbf{9}\pm \textbf{0.9}$\\\hline
  \end{tabular}\label{tab:latents-exp1}
\end{table}

\begin{table}
\centering 
\caption{\glspl{CR-VAE} learn representations that yield higher accuracy on downstream classification than their base \glspl{VAE}. These results correspond to the accuracy from a linear classifier that was fitted on the training. We fed this classifier with the representations learned by each method. On both \textsc{mnist} and \textsc{cifar}-10, \glspl{CR-VAE} yield higher accuracy.}
\begin{tabular}{l c c }
    \toprule
    Method & \textsc{mnist} & \textsc{cifar}-10 \\
    \midrule
   \gls{VAE}  & 98.5 & 32.6 \\
   \gls{VAE}+Aug & 98.9 & 40.1 \\
   \gls{CR-VAE}  & \textbf{99.4} & \textbf{44.7} \\
   \midrule
   \gls{IWAE} & 98.6 & 35.8 \\
   \gls{IWAE}+Aug & \textbf{99.9} & 37.1 \\
   \acrshort{CR-IWAE} & \textbf{99.9} & \textbf{44.8} \\
   \midrule
   $\beta$- \gls{VAE} ($0.5$) & 97.6 & 27.0 \\
   $\beta$- \gls{VAE} ($0.5$)+Aug & 98.7 & 27.6 \\
   $\beta$- \gls{CR-VAE} ($0.5$) &  \textbf{98.9} & \textbf{30.0} \\
   \midrule
   $\beta$- \gls{VAE} ($10$) & 99.4 & 36.5 \\
   $\beta$- \gls{VAE} ($10$)+Aug & \textbf{99.6} & 42.1 \\
   $\beta$- \gls{CR-VAE} ($10$) & \textbf{99.6} & \textbf{46.1} \\
   \bottomrule
\end{tabular}
\label{tab:accuracy}
\end{table}

\begin{table}
\caption{\glspl{CR-VAE} generalize better than their base \glspl{VAE} on almost all cases; they achieve lower negative log-likelihoods. Although training the base \glspl{VAE} with the augmented data improves generalization, adding the consistency regularization term further improves generalization performance.}
  \centering
  \begin{tabular}{l|ccc}
    \hline
    Method & \textsc{mnist} & \textsc{omniglot} & \textsc{celeba} \\
    \hline
    \gls{VAE} & $83.7\pm 0.3$ & $128.2\pm 0.8$ & $66.1\pm 0.2$\\
    \gls{VAE} + Aug & $82.8\pm 0.4$ & $125.7\pm 0.2$ & $66.0\pm 0.2$\\ 
    \gls{CR-VAE} & $\textbf{81.2}\pm \textbf{0.2}$ & $\textbf{124.1}\pm \textbf{0.1}$ & $\textbf{65.9}\pm \textbf{0.2}$\\ \hline
    \gls{IWAE} & $81.7\pm 0.3$ & $127.5\pm 0.5$ & $65.3\pm 0.1$ \\
    \gls{IWAE}+Aug & $80.4\pm 0.2$ & $125.0\pm 0.6$ & $65.3\pm 0.1$ \\
    \acrshort{CR-IWAE} & $\textbf{79.7}\pm \textbf{0.3}$ & $\textbf{123.6}\pm \textbf{0.5}$ & $\textbf{65.0}\pm \textbf{0.2}$ \\\hline
    $\beta$-\gls{VAE} ($0.5$) & $92.6\pm 0.3$ & $137.1\pm 0.2$ & $68.7\pm 0.2$\\
    $\beta$-\gls{VAE} ($0.5$) + Aug & $90.0\pm 0.5$  & $134.6\pm 0.5$ & $68.8\pm 0.2$ \\
    $\beta$-\gls{CR-VAE} ($0.5$) & $\textbf{85.7}\pm \textbf{0.6}$ & $\textbf{132.5}\pm \textbf{0.3}$  & $\textbf{68.2}\pm \textbf{0.1}$ \\\hline
     $\beta$-\gls{VAE} ($10$) & $\textbf{126.1}\pm 1.8$ & $157.5\pm 1.1$ & $92.7\pm 0.5$ \\
    $\beta$-\gls{VAE} ($10$) + Aug & $127.1\pm 1.0$ & $\textbf{157.3}\pm 0.5$ & $92.7\pm 0.3$ \\
    $\beta$-\gls{CR-VAE} ($10$) & $126.2\pm 0.5$ & $157.6\pm 0.6$ & $\textbf{92.6}\pm \textbf{0.1}$ \\\hline
  \end{tabular}\label{tab:nll}
\end{table}

In the second set of experiments we apply our method to a large-scale \gls{VAE}, the latest \gls{NVAE}~\citep{vahdat2020nvae}. We use \textsc{mnist}, \textsc{cifar}-10, and \textsc{celeba} as datasets for this experiment. We increased the resolution for the \textsc{celeba} dataset for this experiment to $64$x$64$. We reach the same conclusions as for the first sets of experiments; \glspl{CR-VAE} improve the learned representations and generalization of their base \glspl{VAE}. In this particular setting, the \gls{CR-NVAE} achieves state-of-the-art generalization performance on both \textsc{mnist} and \textsc{cifar}-10. This state-of-the-art performance couldn't be reach simply by training the \gls{NVAE} with augmentations, as our results show. 

Finally, in a third set of experiments, we apply our regularization technique to a 3D point-cloud dataset called ShapeNet~\citep{shapenet}. We adapt a high-performing auto-encoding method called FoldingNet~\citep{foldingnet} to its \gls{VAE} counterpart and apply the method we described in this paper to that \gls{VAE} variant on the ShapeNet dataset. We found that adding consistency regularization yields better learned representations. 

We next describe in great detail the set up for each of these experiments and the results showcasing the usefulness of the regularization method we propose in this paper. 

\subsection{Application to standard \glspl{VAE} on benchmark datasets}

We apply consistency regularization, as described in this paper, to the original \gls{VAE}, the \gls{IWAE}, and the $\beta$-\gls{VAE}. We now describe the set up and results for this experiment. 

\parhead{Datasets.} We study three benchmark datasets that we briefly describe below. 
We first consider \textsc{mnist}. \textsc{mnist} is a handwritten digit recognition dataset with $60,000$ images in the training set and $10,000$ images in the test set \citep{mnist}. We form a validation set of $10,000$ images randomly sampled from the training set.

We also consider \textsc{omniglot}, a handwritten alphabet recognition dataset \citep{omniglot}. This dataset is composed of $19,280$ images. We use $16,280$ randomly sampled images for training and $1,000$ for validation and the remaining $2,000$ samples for testing. 

Finally we consider \textsc{celeba}. It is a dataset of faces, consisting of $162,770$ images for training, $19,867$ images for validation, and $19,962$ images for testing \citep{celeba}. We set the resolution to $32$x$32$ for this experiment.

\parhead{Transformations $t(\tilde \mbx \vert \mbx)$.} We consider three transformations variants for image data $t(\tilde \mbx \vert \mbx)$. The first randomly translates an image $[-2,2]$ pixels in any direction. The second transformation randomly rotates an image uniformly in $[-15,15]$ degrees clockwise. Finally the third transformation randomly scales an image by a factor uniformly sampled from $[0.9,1.1]$. 
\begin{table}
\centering
\caption{The regularization term $\lambda$ affects both generalization performance and the quality of the learned representations. Many values of $\lambda$ perform better than the base \gls{VAE}. However a large enough value of $\lambda$, e.g. $\lambda = 1$, can lead to worse performance than the base \gls{VAE} because for large values of $\lambda$ the regularization term takes over the data-term in the objective function.}
\begin{tabular}{l l c c c}
    \toprule
     & $\lambda$ & \textsc{MI} & \textsc{AU} & \textsc{NLL}  \\
    \midrule
   \gls{VAE} & $--$ & $124.5$ & $36$ & $83.7$ \\
   \gls{CR-VAE} & $0.001$ & $125.0$ & $38$ & $83.5$ \\
   \gls{CR-VAE} & $0.01$ & $125.9$ & $41$ & $82.4$ \\
   \gls{CR-VAE} & $\textbf{0.1}$ & $\textbf{126.3}$ & $\textbf{47}$ & $\textbf{81.2}$ \\
   \gls{CR-VAE} & $1$ & $124.3$ &	$47$ & $83.9$ \\
   \bottomrule
\end{tabular}
\label{tab:ablation1}
\end{table}

\begin{table}
\centering
\caption{The choice of augmentation affects both generalization performance and the quality of the learned representations. Jointly using all augmentations works best.}
\begin{tabular}{l c c c}
    \toprule
     Augmentation & \textsc{MI} & \textsc{AU} & \textsc{NLL}  \\
    \midrule
   Rotations only & $125.8$ & $45$ & $82.1$ \\
   Translations only & $126.1$ & $45$ & $81.9$ \\
   Scaling only & $125.1$ &	$42$ & $82.7$ \\
   All & $\textbf{126.3}$ & $\textbf{47}$ & $\textbf{81.2}$\\
   \bottomrule
\end{tabular}
\label{tab:ablation2}
\end{table}

\parhead{Evaluation metrics.} The regularization method we propose in this paper is mainly aimed at improving the learned representations of \glspl{VAE}. To assess these representations we use three metrics: mutual information, number of active latent units, and accuracy on a downstream classification task. We also evaluate the effect of the proposed method on generalization to unseen data. For that we also report negative log-likelihood. We define each of these metrics next. 

\emph{Mutual information (MI).} The first quality metric is the mutual information $I(\mbz; \mbx)$ between the observations and 
the latents under the joint distribution induced by the encoder, 
\begin{align}
    I(\mbz; \mbx) &= \mathbb{E}_{p_d(\mbx)} \left[KL(q_{\phi}(\mbz \vert \mbx) \vert\vert p(\mbz)) - \gls{KL}(q_{\phi}(\mbz) || p(\mbz)) \right]
\end{align}
where $p_d(\mbx)$ is the empirical data distribution and $q_{\phi}(\mbz)$ is the \emph{aggregated posterior}, the marginal over $\mbz$ induced by the joint distribution defined by $p_d(\mbx)$ and $q_{\phi}(\mbz \vert \mbx)$. The mutual information is intractable but we can approximate it with Monte Carlo. Higher mutual information 
corresponds to more interpretable latent variables. 

\emph{Number of active latent units (AU).} The second quality metrics we consider is the number of active latent units (AU). It is defined in~\citet{iwae} and measures the ``activity" of a dimension of the latent variables $\mbz$. A latent dimension is ``active" if 
\begin{align}
	Cov_{\mbx}(\mathbb{E}_{\mbu \sim q_\phi(\mbu \vert \mbx)}) > \delta
\end{align}
where $\delta$ is a threshold defined by the user. For our experiments we set $\delta = 0.01$. The higher the number of latent active units, the better the learned representations.

\emph{Accuracy on downstream classification.} This metric is calculated by fitting a given \gls{VAE}, taking the learned representations for each data in the test set and computing the accuracy from the prediction of the labels of the images in that same test set by a classifier fitted on the training set. This metric is only applicable to labelled datasets.  

\emph{Negative log-likelihood.} We use negative held-out log-likelihood to assess generalization. Consider an unseen data $\mbx^*$, its negative held-out log-likelihood under the fitted model is
\begin{align}
	\log p_{\theta}(\mbx^*) &= -\log \left(\mathbb{E}_{q_{\phi}(\mbz \vert \mbx^*)} \left[\frac{p_{\theta}(\mbx^*, \mbz)}{q_{\phi}(\mbz \vert \mbx^*)}\right]\right)
	.
\end{align}
This is intractable and we approximate it using Monte Carlo,
\begin{align}
	\log p_{\theta}(\mbx^*) &\approx -\log \frac{1}{S} \sum_{s=1}^{S}\frac{p_{\theta}(\mbx^*, \mbz^{(s)})}{q_{\phi}(\mbz^{(s)} \vert \mbx^*)}
\end{align}
where $\mbz^{(1)}, \dots, \mbz^{(S)} \sim q_{\phi}(\mbz \vert \mbx^*)$. 

\parhead{Settings.} \label{sec:exp_settings} The \glspl{VAE} are built on the same architecture as \cite{wae}.
The networks are trained with the Adam optimizer with a learning rate of $10^{-4}$ \citep{adam} and trained for $100$ epochs with a batch size of $64$. We set the dimensionality of the latent variables to $50$, therefore the maximum number of active latent units in the latent space is $50$. 
We found $\lambda = 0.1$ to be best according to cross-validation using held-out log-likelihood and exploring the range  $[1e^{-4}, 1.0]$ datasets. In an ablation study we explore $\lambda = 0$. For the $\beta$-\gls{VAE} we set $\lambda = 0.1 \cdot \beta$ and study both $\beta = 0.1$ and $\beta = 10$, two regimes under which the $\beta$-\gls{VAE} performs qualitatively very differently~\citep{higgins2017beta}. All experiments were done on a GPU cluster consisting of Nvidia P100 and RTX. The training took approximately 1 day for most experiments.

\parhead{Results.} \Cref{tab:latents-exp1} shows that on all the three benchmark datasets all the different \gls{VAE} variants we studied, consistency regularization as developed in this paper always improves the quality of the learned representations as measured by mutual information and the number of active latent units. These results are confirmed by the numbers shown in \Cref{tab:accuracy} where \glspl{CR-VAE} always lead to better accuracy on downstream classification.   

We proposed consistency regularization as a way to improve the quality of the learned representations. Incidentally, \Cref{tab:nll} also shows that it can improve generalization as measured by negative log-likelihood. 

\parhead{Ablation Study.} We now look at the impact of each factor that goes into the regularization method we introduced in this paper using \textsc{mnist}. We test the impact of the regularization term $\lambda$ and the impact of the choice of augmentation on all metrics. \Cref{tab:ablation1} and \Cref{tab:ablation2} show the results.

\Cref{tab:ablation1} shows that even small consistency regularization (a small $\lambda$ value) results in improvement over the base \gls{VAE} but that a large enough $\lambda$ value can hurt performance.

\Cref{tab:ablation2} shows that rotations and translations are more important than scaling, but the combination of all three augmentations works best for \glspl{CR-VAE}.

\parhead{Comparison to Contrastive Learning.} We look at how \glspl{CR-VAE} compare against a popular and advanced contrastive-learning-based technique, the \emph{triplet loss} \citep{schroff2015facenet} using \textsc{mnist}. \Cref{tab:contrastive} shows that the \gls{CR-VAE} outperforms the triplet loss on both generalization performance and quality of learned representations. \Cref{tab:contrastive} also confirms existing literature showing simply applying augmentations can outperform complex contrastive learning-based methods such as the triplet loss~\citep{kostrikov2020image, sinha2021s4rl}.
\begin{table}
\centering
\caption{The \gls{CR-VAE} outperforms a popular and advanced contrastive learning technique called \emph{triplet loss} on both generalization performance and quality of learned representations.}
\begin{tabular}{l c c c}
    \toprule
     Method & \textsc{MI} & \textsc{AU} & \textsc{NLL}  \\
    \midrule
   \gls{VAE} & $124.5$ & $36$ &	$83.7$ \\
   \gls{VAE} + augmentations & $125.9$ & $42$ &	$82.8$\\
   \gls{VAE} + triplet loss & $124.9$ & $39$ & $83.1$ \\
   \gls{CR-VAE} & $\textbf{126.3}$ & $\textbf{47}$ & $\textbf{81.2}$ \\
   \bottomrule
\end{tabular}
\label{tab:contrastive}
\end{table}

\begin{table}
\centering
\caption{The \glspl{CR-NVAE} learns better representations than the base \gls{NVAE} as measured by accuracy on a downstream classification on both \textsc{mnist} and \textsc{cifar}-10. We get to this same conclusion when looking at the number of active units as an indicator for the quality of the learned latent representations; \gls{CR-NVAE} recovers $226$ units whereas \gls{NVAE} recovers $211$ units.}
\begin{tabular}{l c c }
    \toprule
    Method & \textsc{mnist} & \textsc{cifar}-10 \\
    \midrule
   \gls{NVAE} & \textbf{99.9} & 57.9 \\
   \gls{NVAE}+Aug & \textbf{99.9} & 66.4 \\
   \acrshort{CR-NVAE} & \textbf{99.9} & \textbf{71.4} \\
   \bottomrule
\end{tabular}
\label{tab:accuracy-nvae}
\end{table}

\begin{table}[t]
\centering
\caption{
Large-scale experiments with \glspl{NVAE} with and without consistency-regularization on 3 benchmark datasets: dynamically binarized \textsc{mnist}, \textsc{cifar}-10 and \textsc{celeba}. We report generalization using negative log-likelihood on \textsc{mnist} and bits per dim on \textsc{cifar}-10 and \textsc{celeba}. On all datasets consistency regularization improves generalization performance. In particular \gls{CR-NVAE} achieves state-of-the-art performance on \textsc{mnist} and \textsc{cifar}-10.}
\begin{tabular}{c c c c c c c}
\toprule
  &  \textsc{mnist} ($28 \times 28$) & \textsc{cifar}-10 ($32 \times 32$) & \textsc{celeba} ($64 \times 64$) \\
\midrule

\gls{NVAE} & 78.19 & 2.91 & 2.03 \\
\gls{NVAE}+Aug & 77.53 & 2.70 & 1.96 \\
\acrshort{CR-NVAE} & \textbf{76.93}  & \textbf{2.51} & \textbf{1.86} \\
\bottomrule
\end{tabular}
\label{tab:results_large_scale}
\end{table}
\subsection{Application to the large-scale \gls{NVAE} on benchmark datasets}

Along with standard VAE variants, we also experiment with a large scale state-of-the-art \gls{VAE}, the \gls{NVAE}\citep{vahdat2020nvae}. Similar to before, we simply add consistency regularization using the image-based augmentations techniques to the NVAE model and experiment on benchmark datasets: \textsc{mnist} \citep{mnist}, \textsc{cifar}-10 \citep{cifar} and \textsc{celeba} \citep{celeba}.

The results for large scale generative modeling are tabulated in \Cref{tab:results_large_scale} and \Cref{tab:accuracy-nvae}, where we see that using \acrshort{CR-NVAE} we are able to learn representations that yield better accuracy on downstream classification and set new state-of-the-art values on each of the datasets, improving upon the baseline log-likelihood values. This shows the ability of consistency regularization to work at scale on challenging generative modeling tasks. 

\subsection{Application to the FoldingNet on 3D point-cloud data}

\begin{table}[t]
\centering
\caption{The FoldingNet yields higher accuracy when paired with consistency regularization on the ShapeNet dataset. The results shown here correspond to a FoldingNet that was trained with augmented data, the same used to apply consistency regularization. As can be seen from these results, enforcing consistency through KL as we do in this paper leads to representations that perform well on a downstream classification. Here the classifier used is a linear SVM. We also report mean reconstruction error through Chamfer distance where the same conclusion holds.}
\begin{tabular}{l c c}
    \toprule
    Method & Accuracy & Reconstruction Loss \\
    \midrule
    Folding Net (Aug) & 82.5\% & 0.0355 \\
    CR-Folding Net & \textbf{84.6\%} & \textbf{0.0327} \\
    \bottomrule
\end{tabular}
\label{tab:3d_results}
\end{table}

\begin{figure}[t!]
    \centering
    \subfigure{\includegraphics[width=0.15\textwidth]{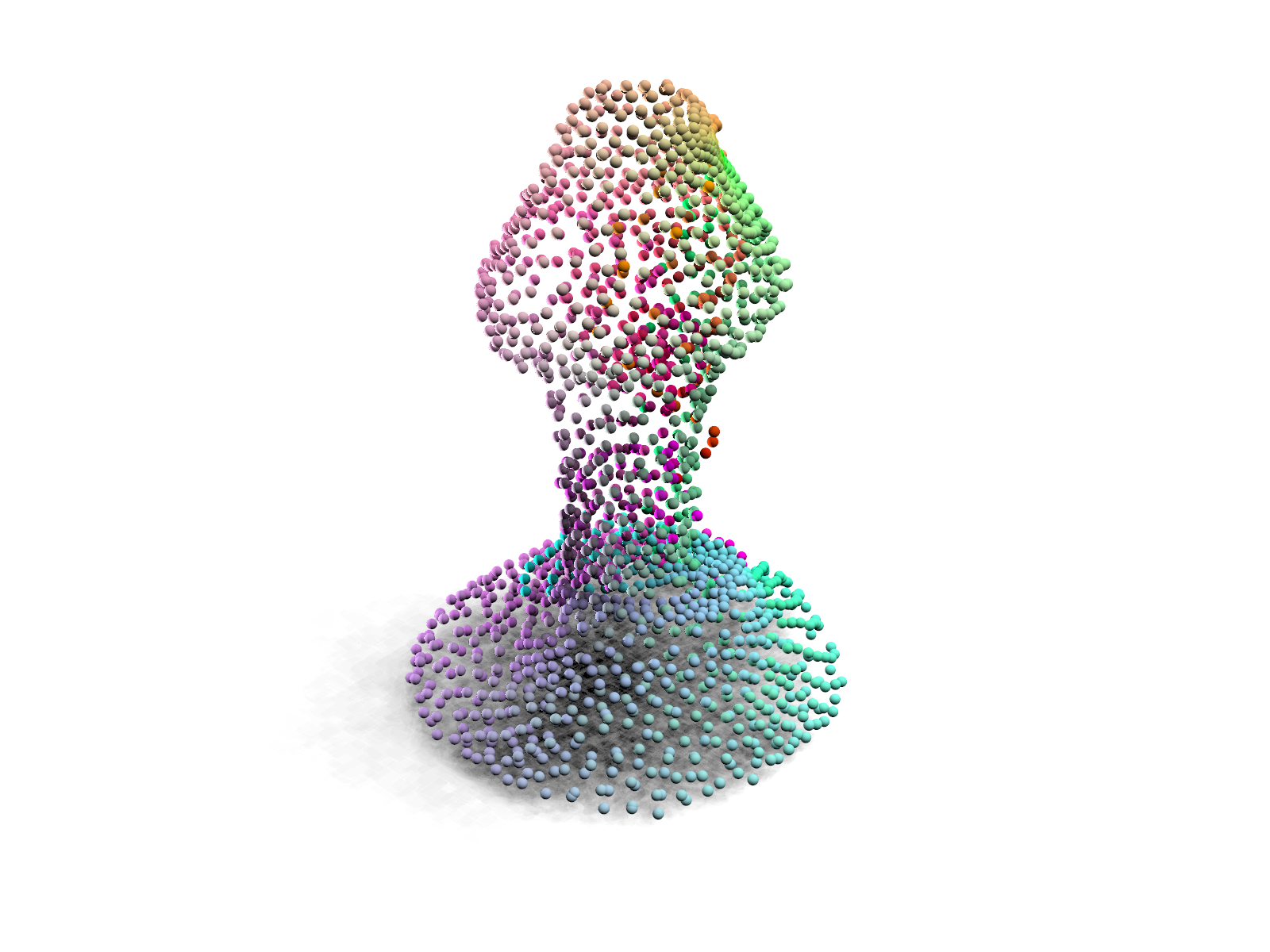}} 
    \subfigure{\includegraphics[width=0.15\textwidth]{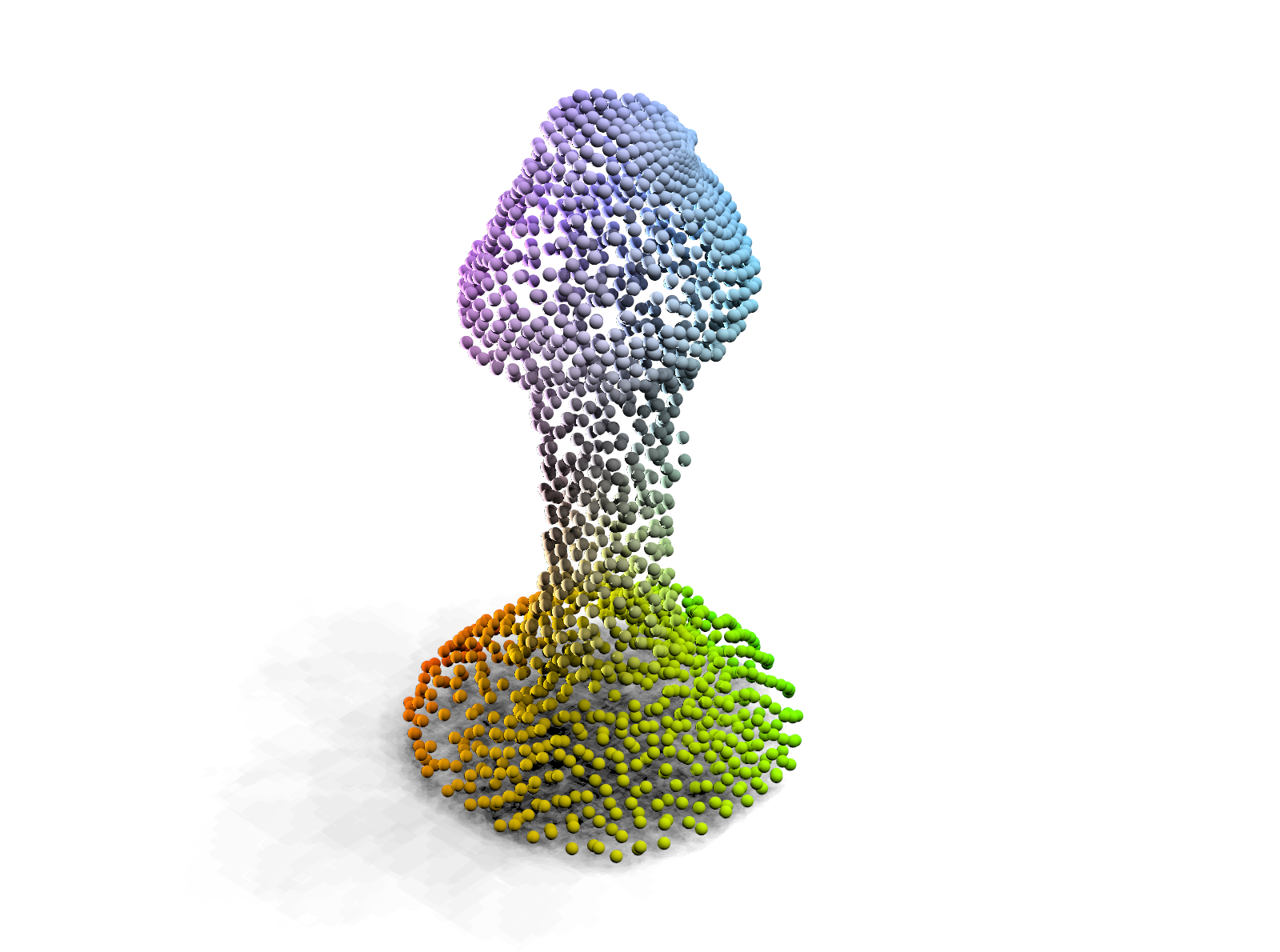}} 
    \subfigure{\includegraphics[width=0.15\textwidth]{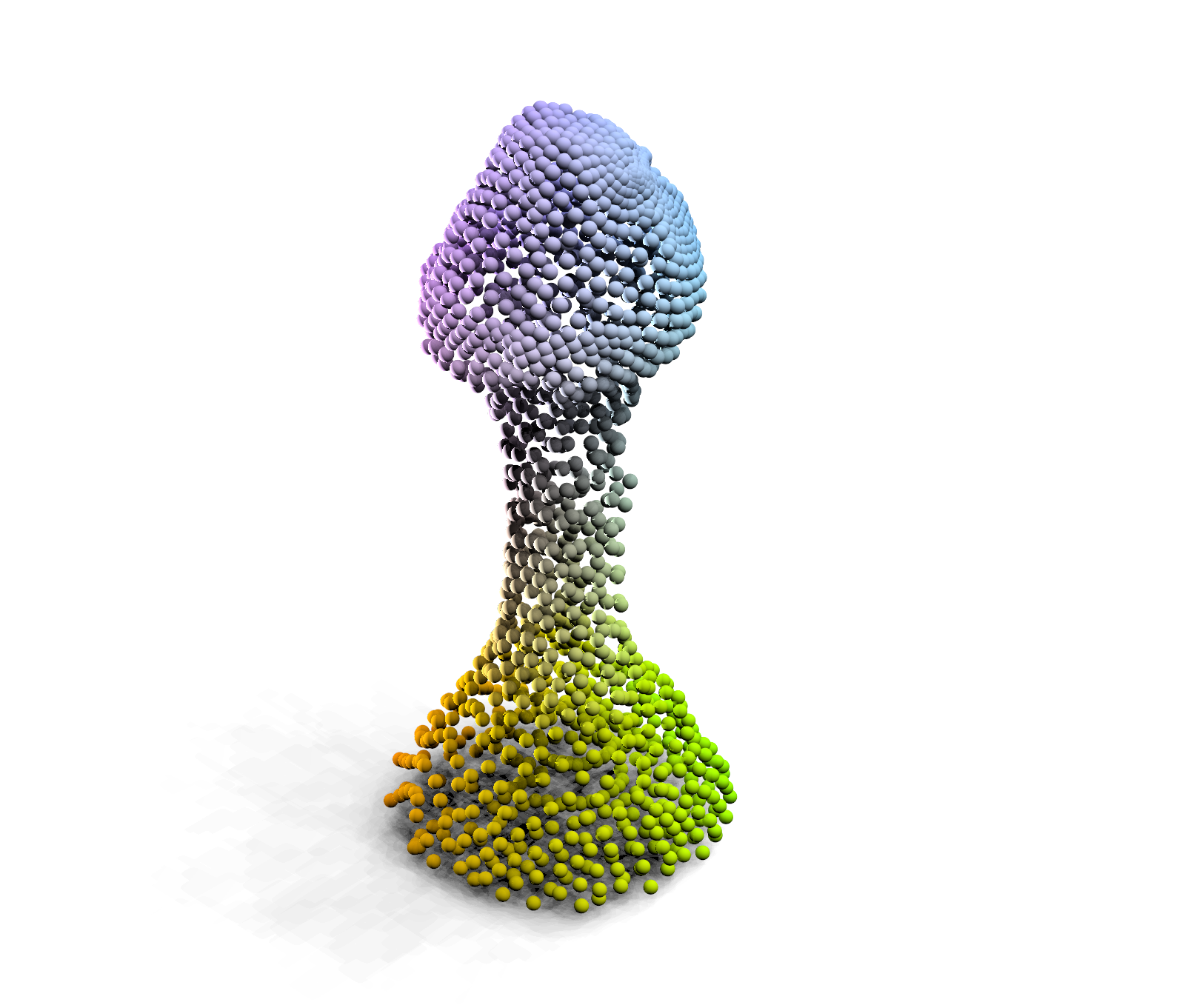}}
    \subfigure{\includegraphics[width=0.15\textwidth]{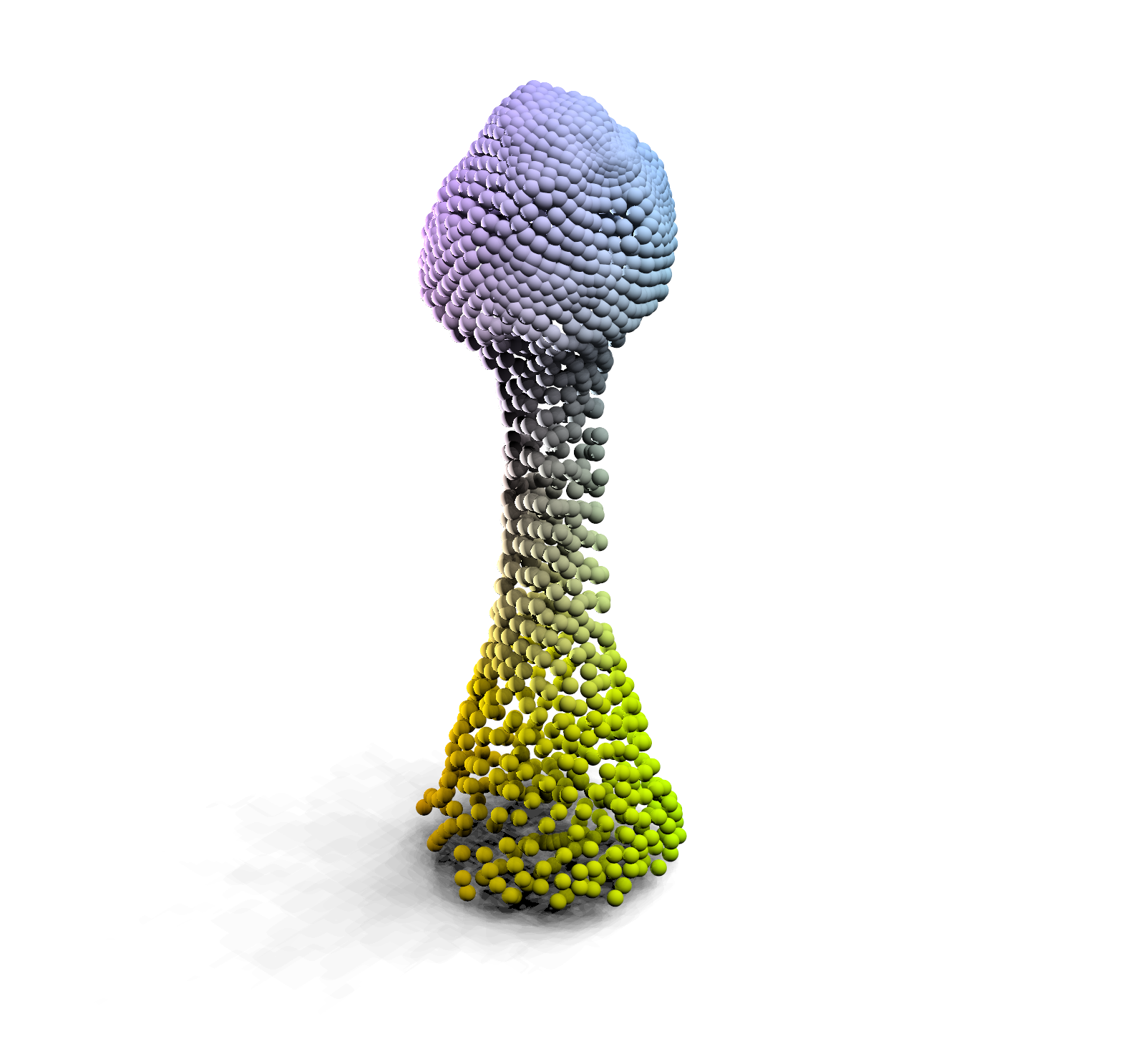}} 
    \subfigure{\includegraphics[width=0.15\textwidth]{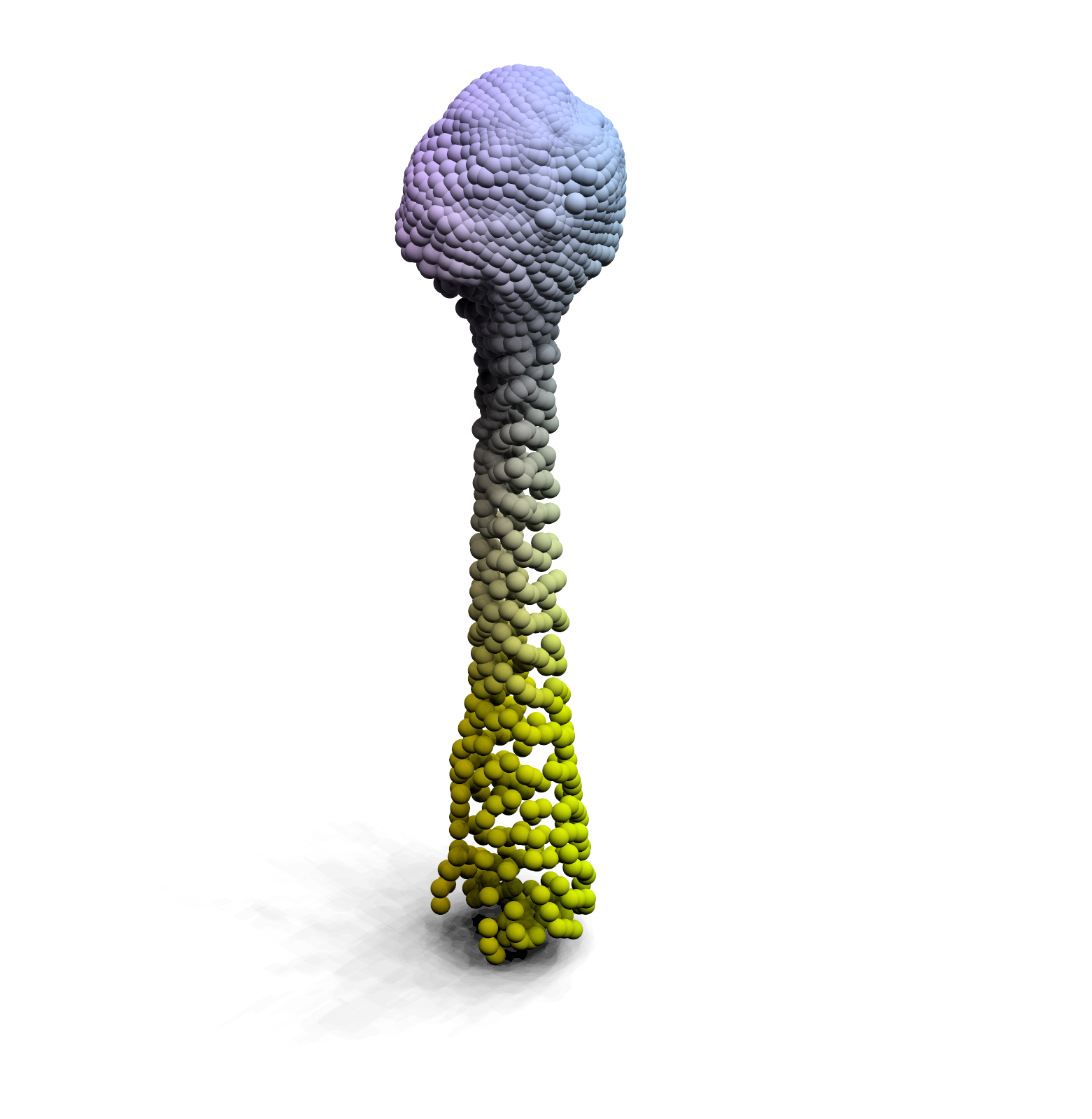}} 
    \subfigure{\includegraphics[width=0.15\textwidth]{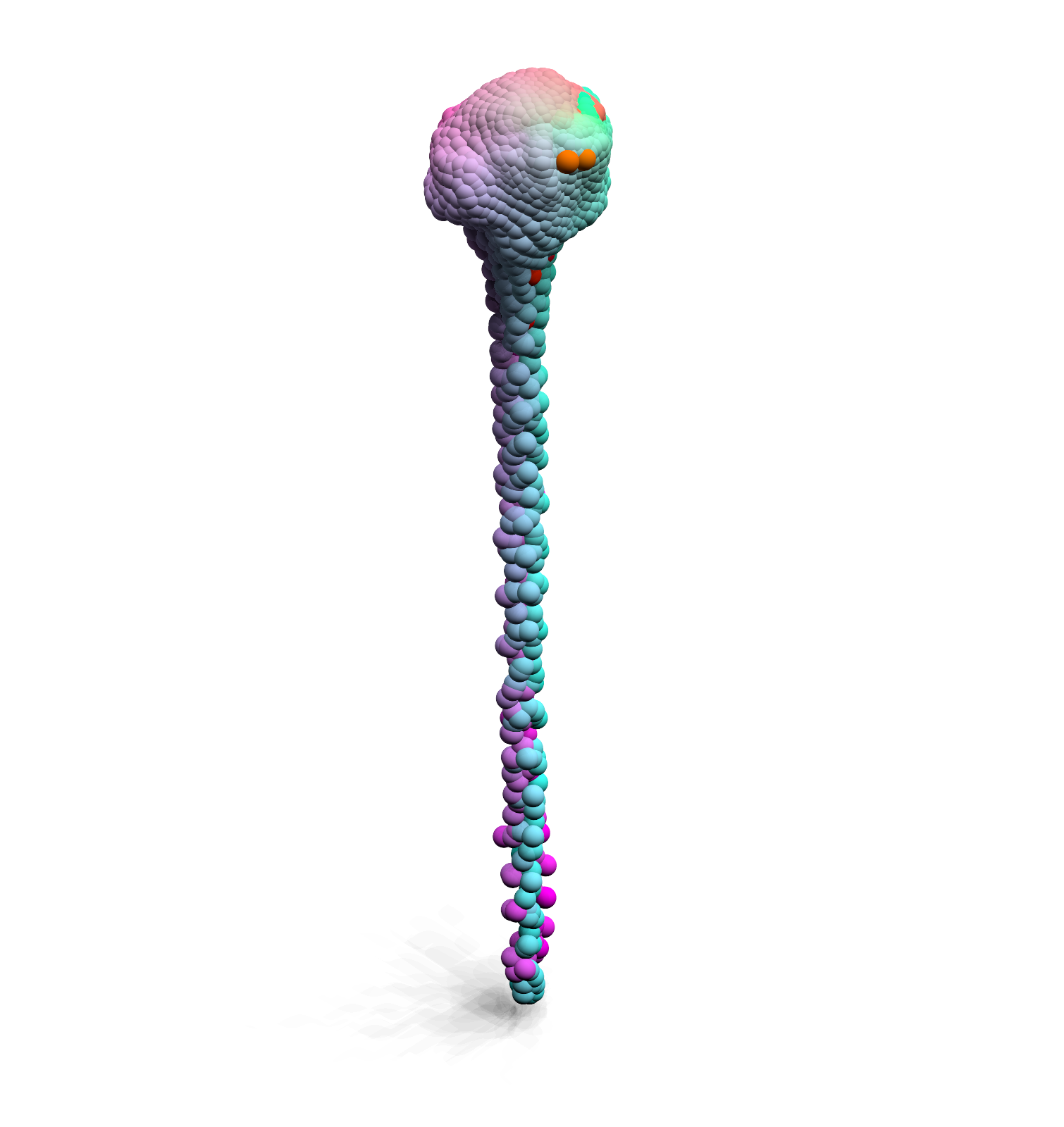}}
    \subfigure{\includegraphics[width=0.15\textwidth]{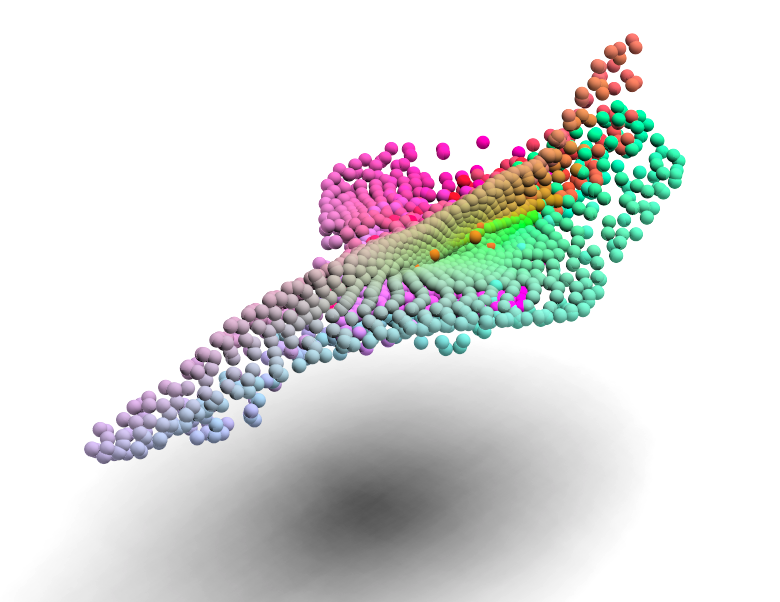}} 
    \subfigure{\includegraphics[width=0.15\textwidth]{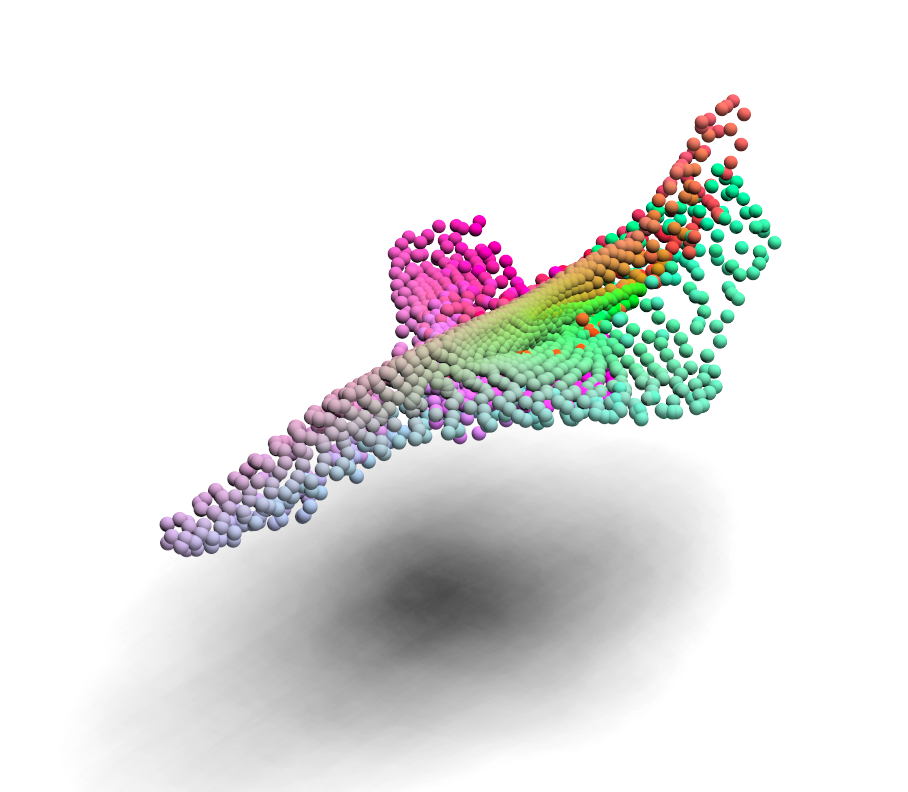}} 
    \subfigure{\includegraphics[width=0.15\textwidth]{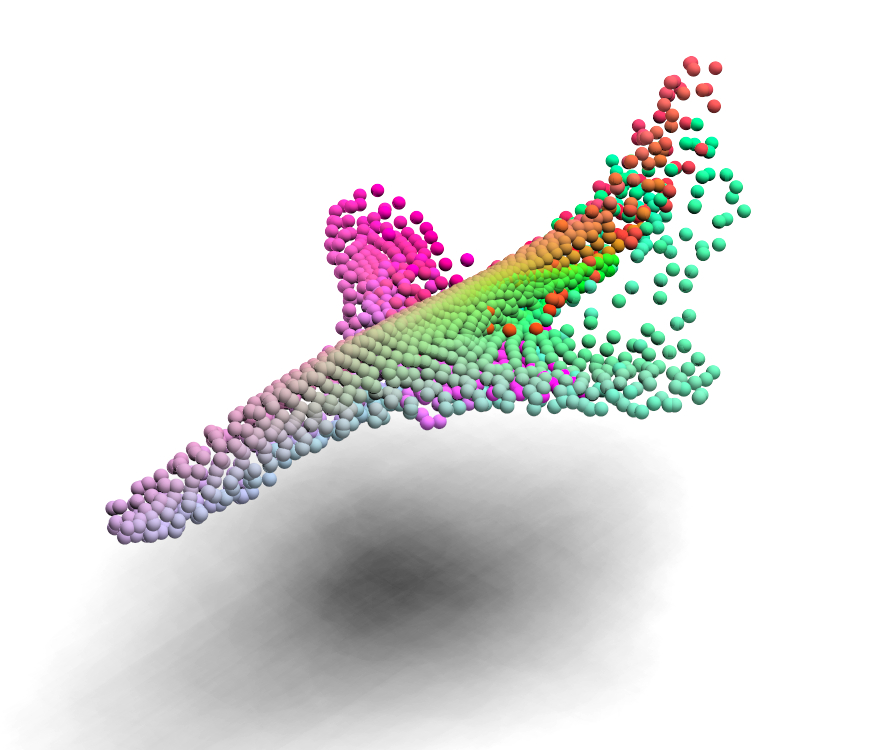}}
    \subfigure{\includegraphics[width=0.15\textwidth]{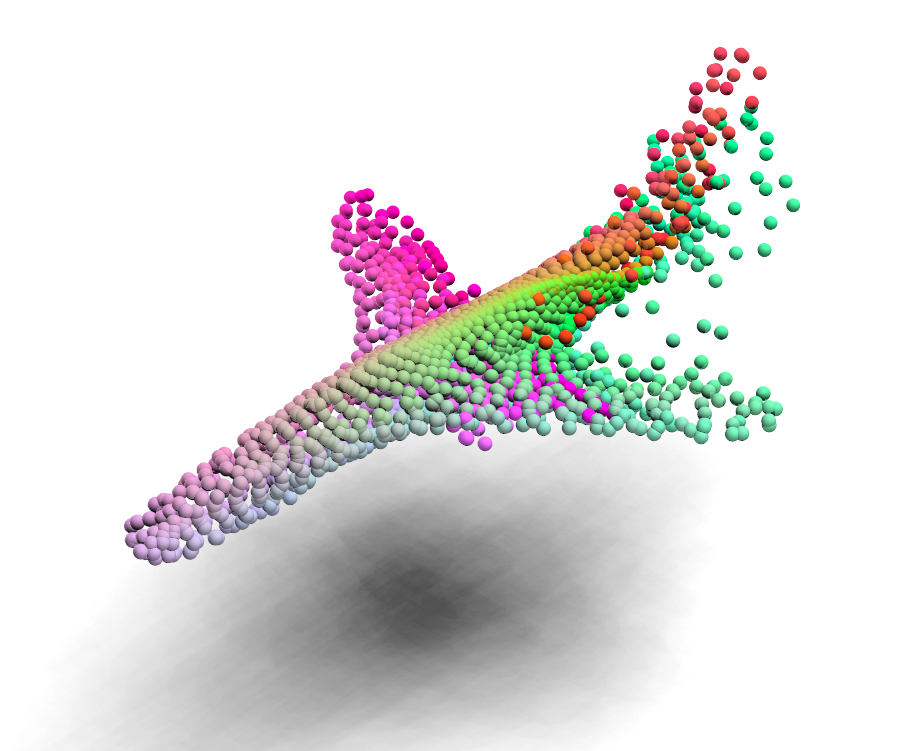}} 
    \subfigure{\includegraphics[width=0.15\textwidth]{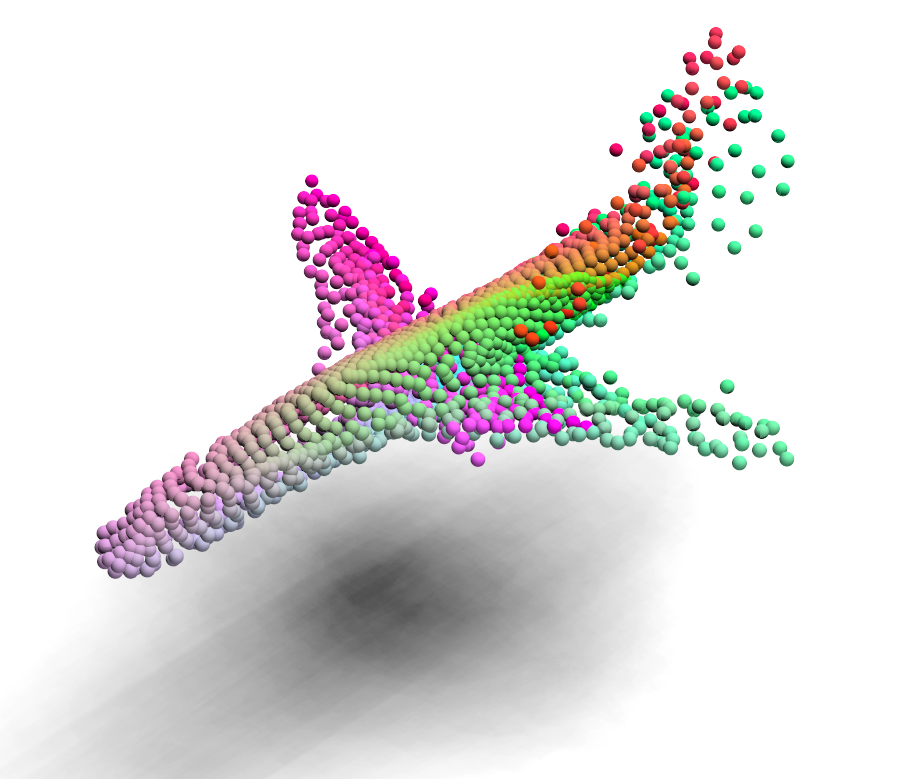}} 
    \subfigure{\includegraphics[width=0.15\textwidth]{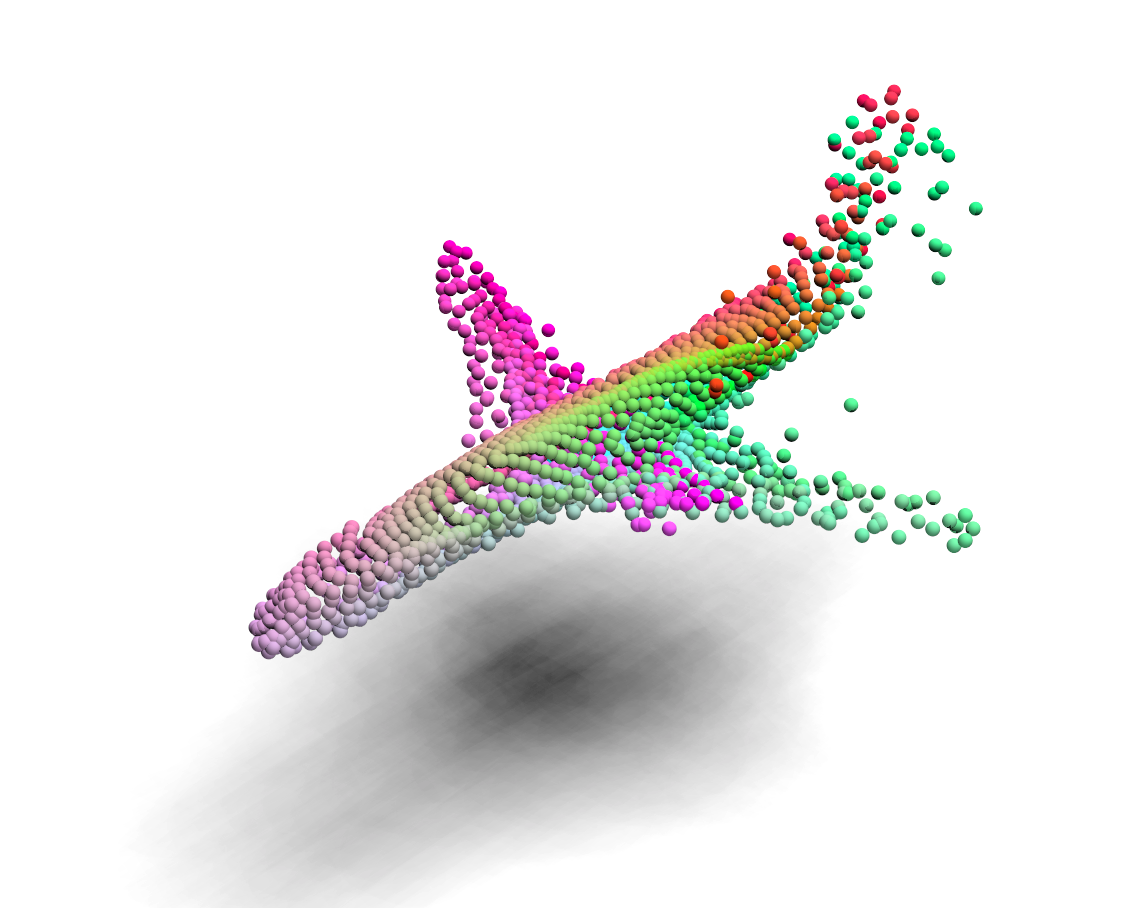}}
    \subfigure{\includegraphics[width=0.15\textwidth]{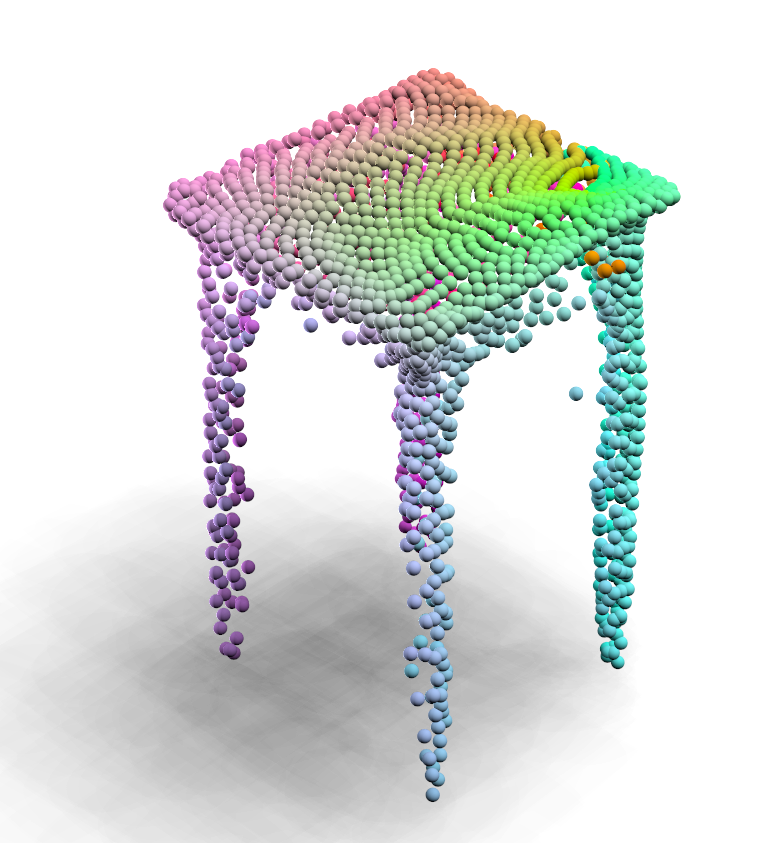}} 
    \subfigure{\includegraphics[width=0.15\textwidth]{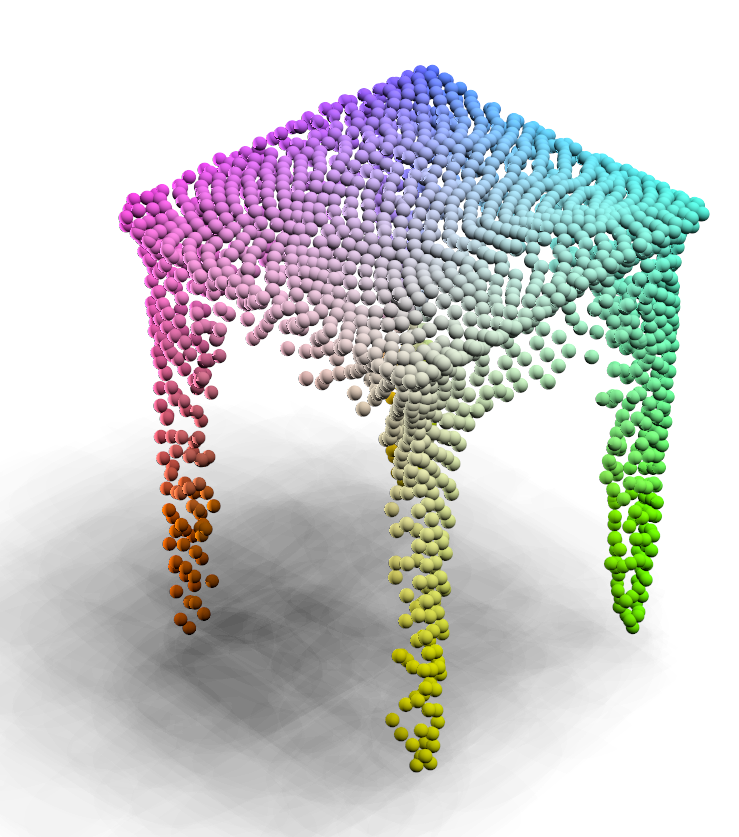}} 
    \subfigure{\includegraphics[width=0.15\textwidth]{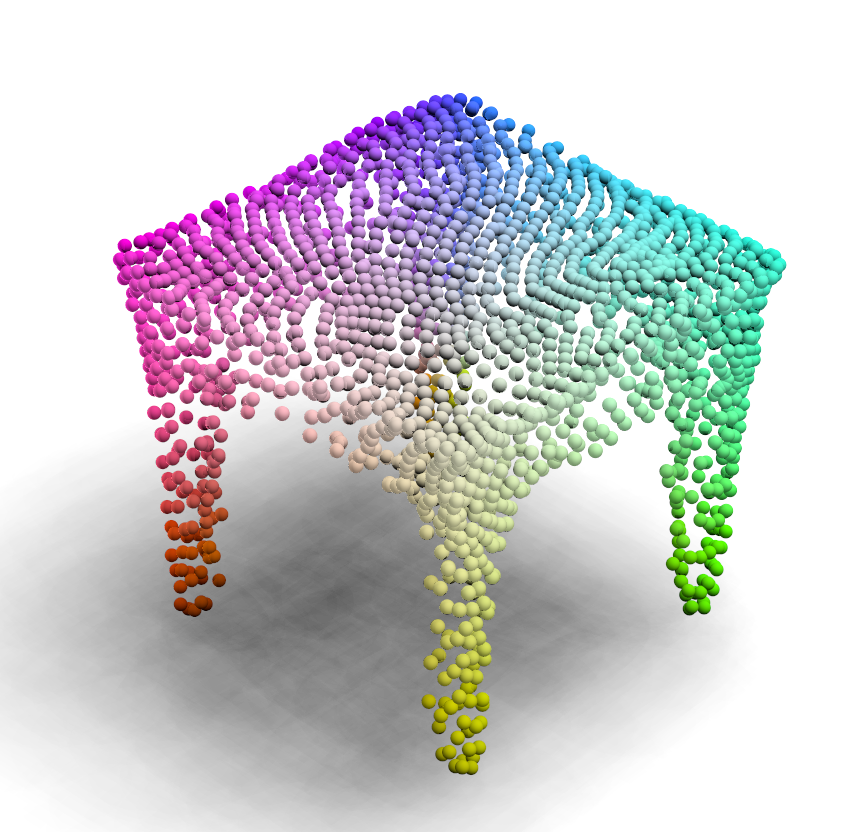}}
    \subfigure{\includegraphics[width=0.15\textwidth]{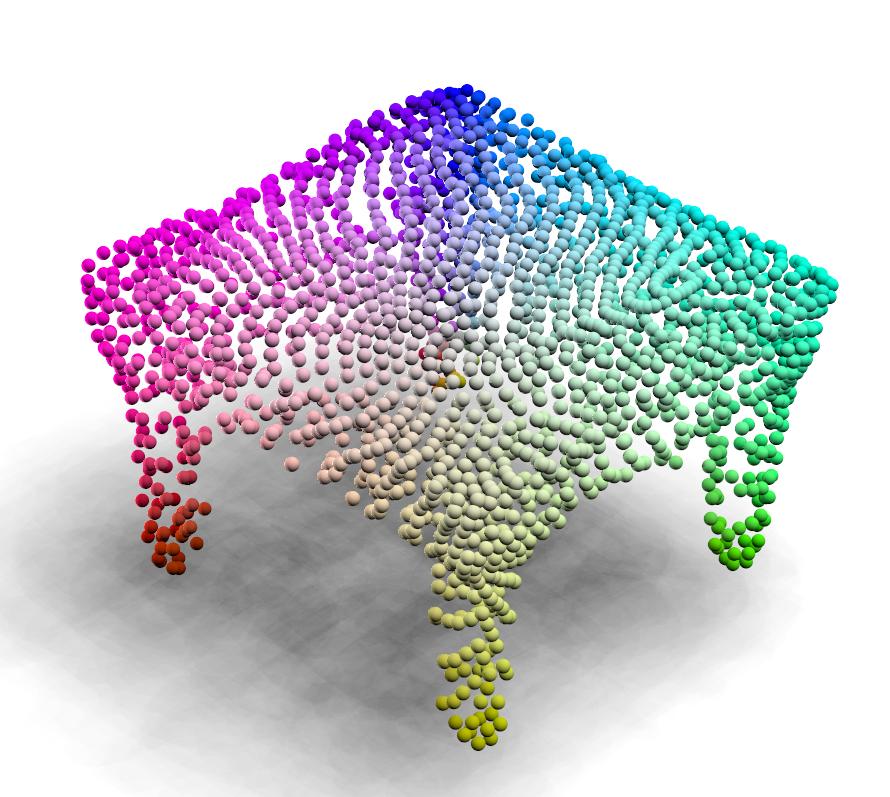}} 
    \subfigure{\includegraphics[width=0.15\textwidth]{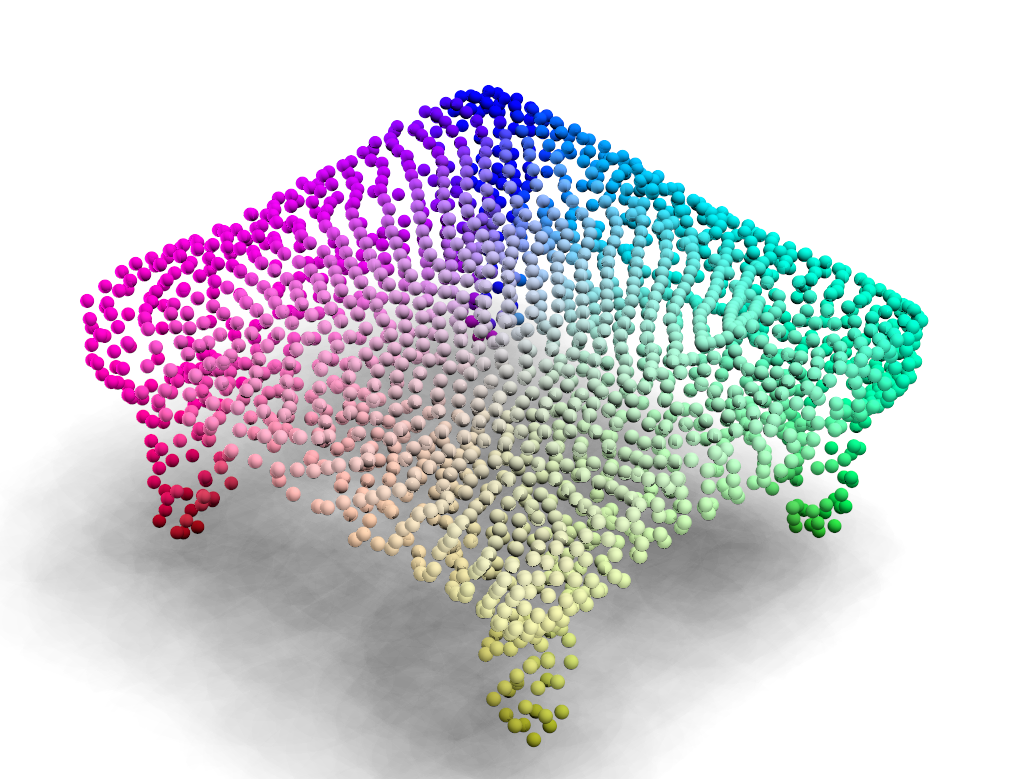}} 
    \subfigure{\includegraphics[width=0.15\textwidth]{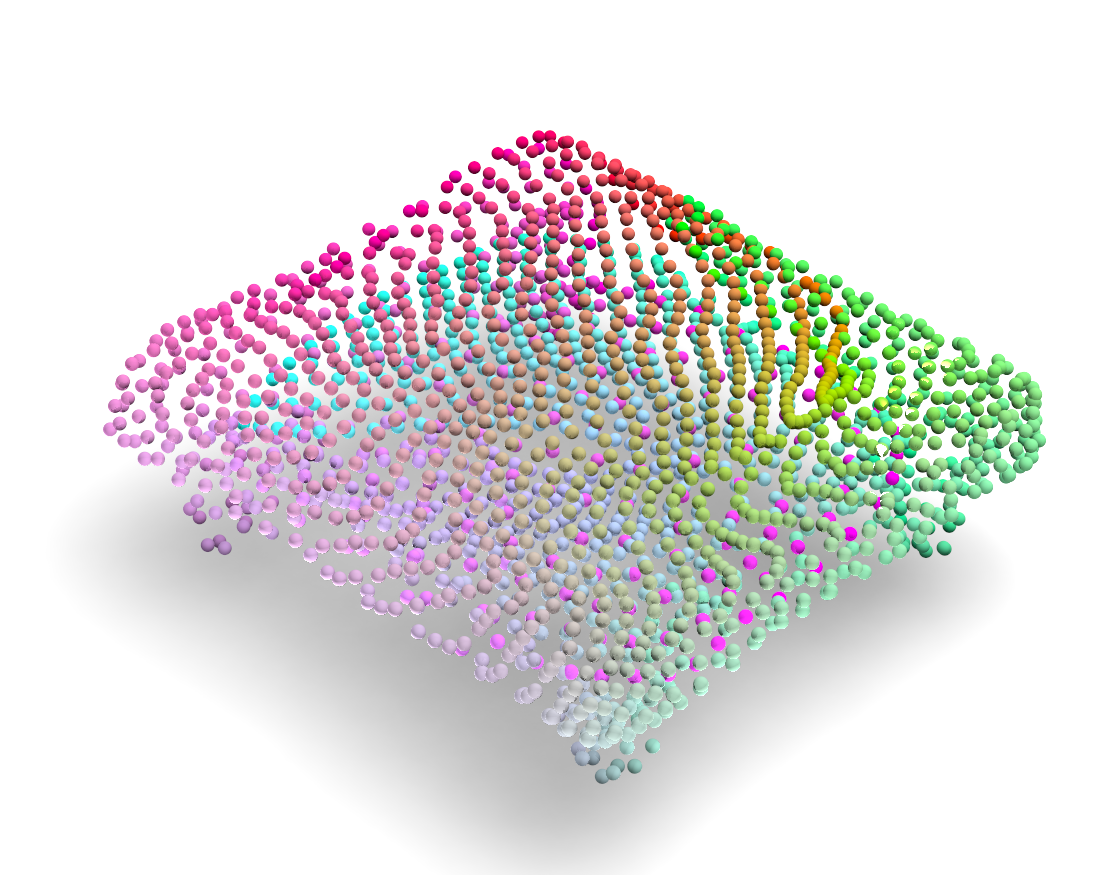}}
    \caption{Interpolation between two samples of a lamp, airplane and table using a trained CR-FoldingNet trained on the ShapeNet dataset. The CR-FoldingNet is able to learn an interpretable latent space.}
    \label{fig:interpolation}
\end{figure}

Along with working with image data, we additionally experiment with 3D point cloud data using a
FoldingNet \cite{foldingnet} and the ShapeNet dataset \cite{shapenet} which consists of 55 distinct
object classes. FoldingNet learns a deep AutoEncoder to learn unsupervised representations from the point cloud data.
To add consistency regularization, we first substitute the AutoEncoder to a \gls{VAE} by adding the KL term from the ELBO to the baseline FoldingNet. We then add the additional consistency regularization KL term to the latent space of FoldingNet.

For the ShapeNet point cloud data, we perform data augmentation using a similar scheme to what we did for the previous experiments, we randomly translate, rotate and add jitter to the $(x,y,z)$ coordinates of the point cloud data. We follow the same scheme detailed in FoldingNet \citep{foldingnet}.

We train both the FoldingNet turned in a \gls{VAE} and the CR-FoldingNet with these augmentations. 
To train CR-FoldingNet, we additionally apply the consistency regularization term as proposed in \Cref{eq:crvae}.
The results on the validation set for reconstruction (as measured by Chamfer distance) and accuracy are shown in \Cref{tab:3d_results}. 

We also visualize the point clouds reconstructions and interpolations between 3 different object classes using a CR-FoldingNet in \Cref{fig:interpolation}. We perform 4 interpolation steps for each of the objects, to highlight the interpretable learned latent space. Additionally, we perform the same interpolation on the baseline FoldingNet model. We show these interpolations in the appendix. 

%% file: sec_discussion.tex

\section{Conclusion}
\label{sec:conclusion}

We proposed a simple regularization technique to constrain encoders of \glspl{VAE} to learn similar latent representations for an image and a semantics-preserving transformation of the image. The idea consists in maximizing the likelihood of the pair of images while  minimizing the \gls{KL} divergence between the variational distribution induced by the encoder when conditioning on the image on one hand, and its transformation, on the other hand. We applied this technique to several \gls{VAE} variants on several  datasets, including a 3D dataset. We found it always leads to better learned representations and also better generalization to unseen data. In particular, when applied to the \gls{NVAE}, the regularization technique we developed in this paper yields state-of-the-art results on \textsc{mnist} and \textsc{cifar}-10.